%% file: arxiv_iclr.tex
\documentclass{article}

\usepackage{arxiv}

\usepackage[utf8]{inputenc} 
\usepackage[T1]{fontenc}    
\usepackage{hyperref}       
\usepackage{url}            
\usepackage{booktabs}       
\usepackage{amsfonts}       
\usepackage{nicefrac}       
\usepackage{microtype}      
\usepackage{graphicx}
\usepackage{natbib}
\usepackage{doi}
\usepackage{float}          

\input{math_commands.tex}

\usepackage[T1]{fontenc}    
\usepackage{url}            
\usepackage{booktabs}       
\usepackage{amsfonts}       
\usepackage{amsmath}        
\usepackage{amssymb}        
\usepackage{nicefrac}       
\usepackage{microtype}      
\usepackage{xcolor}         
\usepackage{colortbl}       
\usepackage{wrapfig}
\usepackage{graphicx}
\usepackage{multirow}
\usepackage{makecell}
\usepackage{subcaption}
\usepackage{comment}

\usepackage{color}

\usepackage{enumitem}
\setlist{nosep,leftmargin=1.4em}
\setlist[itemize]{topsep=2pt,itemsep=1pt,parsep=0pt}
\setlist[enumerate]{topsep=2pt,itemsep=1pt,parsep=0pt}

\newcommand{\ablationsub}[1]{%
  \textit{#1}\enspace%
}

\usepackage[most]{tcolorbox}
\definecolor{dsBox}{HTML}{0F766E}  
\newtcolorbox{dsetexample}[2]{%
  enhanced, breakable,
  colback=dsBox!4, colframe=dsBox,
  colbacktitle=dsBox, coltitle=white,
  fonttitle=\sffamily\bfseries\small,
  title={\textsc{#1}\,\textperiodcentered\, #2},
  boxrule=0.4pt, arc=2pt,
  left=6pt, right=6pt, top=4pt, bottom=4pt,
  before skip=4pt, after skip=4pt,
  fontupper=\small,
}

\title{Hyperbolic Time Series Forecasting with Learnable Decomposition and Manifold Dynamics}

\date{}

\author{
  Sarang Rajendra Patil\thanks{Equal contribution.} \\
  Department of Data Science\\
  New Jersey Institute of Technology\\
  Newark, NJ 07102, USA \\
  \texttt{sp3463@njit.edu} \\
  \And
  Vishal Fenn\footnotemark[1] \\
  Department of Data Science\\
  New Jersey Institute of Technology\\
  Newark, NJ 07102, USA \\
  \texttt{vkf@njit.edu} \\
  \And
  Mengjia Xu\thanks{Corresponding author.} \\
  Department of Data Science\\
  New Jersey Institute of Technology\\
  Newark, NJ 07102, USA \\
  \texttt{mx6@njit.edu} \\
}

\renewcommand{\headeright}{}
\renewcommand{\undertitle}{}

\hypersetup{
pdftitle={Hyperbolic Time Series Forecasting with Learnable Decomposition and Manifold Dynamics},
pdfsubject={cs.LG},
pdfauthor={Sarang Rajendra Patil, Vishal Fenn, Mengjia Xu},
pdfkeywords={Time series forecasting, Hyperbolic geometry, Geometric representation learning},
}

\begin{document}
\maketitle
\setcounter{footnote}{0} 

\begin{abstract}
Multiscale structure is central to time series forecasting: temporal signals contain slowly varying trends, intermediate seasonal patterns, and high-frequency residuals. Traditional forecasters model this multiscale structure in Euclidean space, without representing the ordering it induces among components. We identify that cascaded decomposition orders components by their characteristic frequencies, forming a natural frequency hierarchy. This hierarchy aligns with the exponential volume growth of hyperbolic space, which provides increasing representational capacity with radius and naturally accommodates nested hierarchies. We propose \textbf{HyperForecast}, the first long-term time series forecasting (LTSF) framework operating in hyperbolic space. HyperForecast embeds decomposed temporal segments on the Poincar\'e ball where radial distance captures this nested structure, from low-frequency components near the origin to high-frequency components near the boundary. We model geodesic velocity through two manifold-dynamics variants: (a) autoregressive forecasting via Poincar\'e residual dynamics, and (b) non-autoregressive forecasting that predicts all future segments in parallel. HyperForecast consistently outperforms existing and state-of-the-art baselines on 9 of 11 long-term forecasting benchmarks, achieving up to 58\% lower mean squared error at long horizons. Further analysis reveals that the learned radial embeddings recover the frequency hierarchy and that hyperbolic geometry contributes substantially to forecasting performance. 
~\footnote{Code available at \url{https://anonymous.4open.science/r/Hyperbolic_TimeSeries-3876/}.}
\end{abstract}


\section{Introduction}
\label{sec:introduction}

Time series forecasting is fundamental across domains, from energy management to financial markets. However, capturing multi-scale temporal structures remains a core challenge, since trends evolve slowly, seasonal patterns recur at multiple frequencies, and residuals capture high-frequency stochastic variation. Most existing works \citep{wu2021autoformer, huang2025timebase, niu2026phaseformer, lin2024sparsetsf, xu2024fits, zhou2021informer, zhou2022fedformer, wang2024timemixer, zeng2023transformers, liu2024itransformer, nie2023a, wu2023timesnet} operate in Euclidean space,
lacking an explicit geometric inductive bias for the underlying frequency hierarchy.
We argue that hyperbolic geometry provides a natural manifold for modeling the nested frequency structure induced by time series decomposition.

\begin{figure*}[h]
\begingroup\setlength{\abovecaptionskip}{2pt}\setlength{\belowcaptionskip}{1pt}
    \centering
    \includegraphics[width=0.85\textwidth]{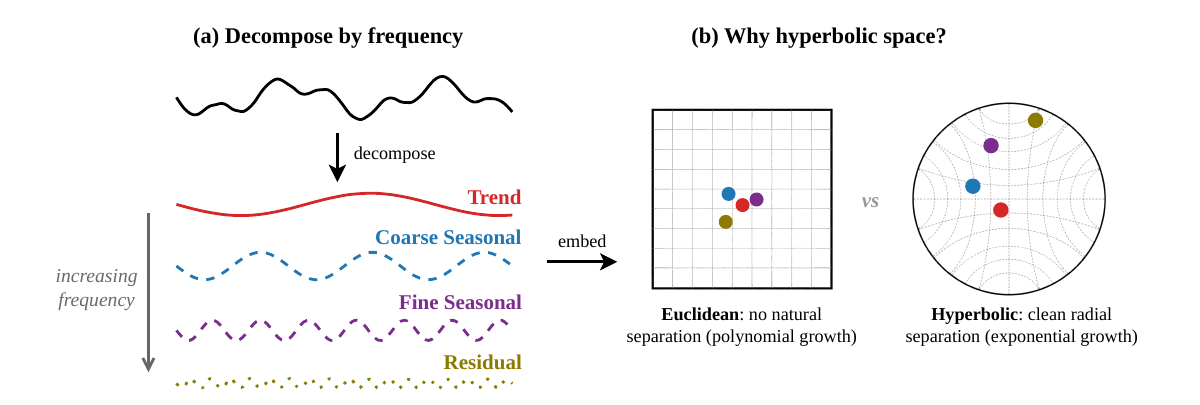}
    \caption{Multiscale temporal decomposition and geometric embedding in HyperForecast. (a) A time series is decomposed into trend, coarse seasonal, fine seasonal, and residual components with increasing frequency. (b) Euclidean space provides no natural separation of these components, whereas the Poincar\'e ball naturally separates them radially through its exponential volume growth.}
    \label{fig:intro}
\endgroup
\end{figure*}

Many existing forecasting methods \citep{wang2024timemixer, zeng2023transformers, bandara2025mstl,wu2021autoformer} decompose time series into separate components, but treat them as geometrically equivalent vectors in flat Euclidean space, thereby ignoring the intrinsic frequency-based ordering induced by the decomposition. This geometric agnosticism constitutes a structural limitation, as the cascaded decomposition naturally induces an ordering: $\text{trend} \prec  \text{seasonal}_{\text{coarse}}  \prec  \text{seasonal}_{\text{fine}} \prec \text{residual}$, where each component captures variation not explained by slower-varying components above it. This ordering defines a frequency-based hierarchy, corresponding to a tree-like containment structure, and hyperbolic geometry has been used to represent such tree-like/hierarchical structures at provably lower distortion than Euclidean space of equal or higher dimension~\citep{nickel2017poincare, ganea2018hyperbolic}. Unlike language or graph data, where such structure is explicit, this ordering emerges implicitly from the temporal dynamics, with slow-varying trends acting as parent components and high-frequency residuals as their descendants. Euclidean geometry, with its polynomial volume growth, provides no natural way to encode this ordering without explicit supervision. Hyperbolic space offers a principled alternative where its constant negative curvature induces exponential volume growth with radius, providing a geometric structure where radial distance naturally encodes frequency-based ordering, as illustrated in Figure~\ref{fig:intro}. Hyperbolic geometry has demonstrated advantages across graph learning, computer vision, and natural language processing~\citep{ganea2018hyperbolic, nickel2017poincare, chami2019hyperbolic, van_Spengler_2023_ICCV, yang2024hypformer, he2026helm, zhang2026hmamba, patil2026hierarchical, he2024language}. Despite this progress, hyperbolic geometry for time series forecasting remains largely unexplored. 

HyperForecast fills this gap by combining (i) learnable decomposition with component-specific convolutional kernels, (ii) segment-level embeddings in Poincar\'e space where radial distance encodes frequency ordering, and (iii) two complementary forecasting paradigms operating via geodesic velocity on the manifold. The autoregressive variant estimates velocity from an exponentially weighted moving window and applies Poincar\'e residual updates for stable sequential rollout, while the non-autoregressive variant aggregates velocity over the full historical lookback and applies a velocity dampener for numerically stable, parallel prediction of all future segments. To the best of our knowledge, HyperForecast is the first long-term time series forecasting (LTSF) architecture to perform forecasting entirely in hyperbolic space, treating decomposed temporal segments as trajectories on a Riemannian manifold rather than points in a flat Euclidean space, a distinction that enables forecasting as geodesic evolution rather than vector extrapolation. We substantiate this argument empirically by computing the Gromov $\delta$-hyperbolicity of the embeddings across all 11 benchmark datasets, indicating the learned representations are broadly tree-like, and this is corroborated by spectral-centroid and radial-ordering analysis of the raw signal itself. 
The key contributions are:

\begin{itemize}
    \item \textbf{First fully hyperbolic forecaster:} We propose, to our knowledge, the first long-term time series forecaster operating in hyperbolic space, applying hyperbolic operations such as exponential maps, M\"obius addition, and Poincar\'e fully-connected (FC) layers.
    \item \textbf{Forecasting via geodesic velocity:} Rather than extrapolating future states as displacement vectors in a flat latent space, we cast forecasting as evolution along a geodesic on the curved manifold itself rather than in a Euclidean tangent space.
    \item \textbf{Dual hyperbolic forecasting paradigms:} We introduce autoregressive (AR) and non-autoregressive (NAR) variants, both forecasting via geodesic velocity on the Poincar\'e manifold but differing in velocity estimation and prediction scheme.
\end{itemize}



\section{Related Works}
\label{sec:related}
Our work sits at the intersection of decomposition-based time series forecasting and hyperbolic representation learning. We review recent advances in both domains and position HyperForecast as the first to combine learnable decomposition with velocity-driven forecasting in hyperbolic space.

PatchTST \citep{nie2023a} segments time series into subseries-level patches with channel independence, improving significantly over full-sequence Transformer models. TimeBase \citep{huang2025timebase} similarly partitions histories into non-overlapping chunks and learns orthogonal basis components for forecasting, while PhaseFormer \citep{niu2026phaseformer} departs from patch-based tokenization altogether, arguing that phase tokens aligned to a series' periodic structure are globally stationary, unlike patch tokens. These segment- or phase-level approaches preserve multi-scale dynamics while improving scalability; HyperForecast aligns with this trend but differs fundamentally, since its segment representations live in hyperbolic manifolds where radial distance encodes frequency hierarchy, rather than in Euclidean spaces.

\citet{papaioannou2022time} proposed an embed-forecast-lift framework using Diffusion Maps and LLE to reduce high-dimensional time series to low-dimensional manifolds before modeling dynamics; DeepEDM \citep{majeedi2025lets} similarly argues that existing methods ignore this underlying structure and instead proposes dynamical systems modeling via Takens' theorem. Both approaches discover manifold geometry empirically from data, whereas HyperForecast argues that the frequency-based hierarchy induced by cascaded decomposition theoretically warrants hyperbolic geometry.

Hyperbolic geometry has proven advantageous for data with nested or tree-like structure in graph learning \citep{chami2019hyperbolic}, NLP \citep{he2026helm, yang2026hyperbolic}, and sequential modeling \citep{zhang2026hmamba, yang2024hypformer}, with recent surveys \citep{peng2021hyperbolic, patil2025hyperbolic} taxonomizing the field, though none address long-horizon forecasting. HyperForecast fills this gap as, to our knowledge, the first complete LTSF framework operating entirely in hyperbolic space, combining frequency-based hierarchy encoding via radial distance, Poincar\'e FC layers for geometric consistency \citep{shimizu2021hyperbolic, van_Spengler_2023_ICCV}, and both AR and NAR forecasting paradigms.

\section{Preliminaries} 
\label{sec:preliminaries}
\paragraph{Hyperbolic manifold (Poincar\'{e} ball).}
The $d$-dimensional Poincar\'{e} ball of constant negative curvature $\mathcal{K} = -c < 0$ is defined as
$\mathbb{P}^d_c = \{\, \mathbf{x} \in \mathbb{R}^d : c\lVert \mathbf{x} \rVert^2 < 1 \,\}$. This is equipped with the Riemannian metric $g_{\mathbf{x}} = \lambda_{\mathbf{x}}^2\, g^{\mathrm{Euc}}$, where $\lambda_{\mathbf{x}} = \frac{2}{1 - c\lVert \mathbf{x} \rVert^2}$ is the conformal factor. As $\mathbf{x}$ approaches the boundary $\lVert \mathbf{x} \rVert \to c^{-1/2}$, $\lambda_{\mathbf{x}}$ diverges, inducing an exponential expansion of geodesic distances that endows $\mathbb{P}^d_c$ with the capacity to embed hierarchical structure at provably lower distortion than any Euclidean space of equal or higher dimension~\citep{nickel2017poincare, ganea2018hyperbolic}.

\paragraph{Basic hyperbolic operations.}Let $\mathcal{M} = \mathbb{P}^d_c$ denote the hyperbolic manifold. The exponential map $\exp^c_{\mathbf{x}} : T_{\mathbf{x}}\mathcal{M} \to \mathcal{M}$ projects a tangent vector $\mathbf{v} \in T_{\mathbf{x}}\mathcal{M}$ onto the manifold by following the geodesic from $\mathbf{x}$ with initial velocity $\mathbf{v}$. The logarithmic map $\log^c_{\mathbf{x}} : \mathcal{M} \to T_{\mathbf{x}}\mathcal{M}$ is its inverse and recovers the corresponding tangent vector. Together, they induce a local diffeomorphism between $\mathbb{P}^d_c$ and $T_{\mathbf{x}}\mathcal{M}$, enabling Euclidean computations to lift and project back from the manifold. At origin $\mathbf{o} = \mathbf{0}$,
\begin{equation} 
\exp_{\mathbf{o}}^{c}(\mathbf{v}) 
= \tanh\!\left(\sqrt{c}\,\lVert \mathbf{v} \rVert\right)
   \frac{\mathbf{v}}{\sqrt{c}\,\lVert \mathbf{v} \rVert + \varepsilon}, \quad
\log_{\mathbf{o}}^{c}(\mathbf{x}) 
= \tanh^{-1}\!\left(\sqrt{c}\,\lVert \mathbf{x} \rVert\right)
   \frac{\mathbf{x}}{\sqrt{c}\,\lVert \mathbf{x} \rVert + \varepsilon},
\label{eq:log_map}
\end{equation}

where $\varepsilon>0$ is a small constant added throughout for numerical stability. The hyperbolic distance $d_c(\mathbf{o}, \mathbf{x})$ from the origin $\mathbf{o}$ determines the radial position of $\mathbf{x}$ on $\mathbb{P}^d_c$, inducing a natural frequency stratification, where points near the origin encode low-frequency components (trend, coarse seasonal) and points near the boundary encode high-frequency components (fine seasonal, residual). This geometric alignment between radial depth and temporal scale is the key structural property that distinguishes $\mathbb{P}^d_c$ from Euclidean spaces. Full definitions for M\"obius addition $\oplus$ and scalar multiplication $\otimes_c$, the general exponential and logarithmic maps, hyperbolic distance, and parallel transport are provided in Appendix~\ref{app:hyperbolic-defs}.

\paragraph{Poincar\'e FC layer.}
A Poincar\'e fully-connected (FC) layer with weight $\mathbf{W} \in \mathbb{R}^{d\times d}$ and bias $\mathbf{b}\in\mathbb{R}^d$ extends the Euclidean affine map to $\mathcal{M}$ via the M\"obius matrix-vector product plus bias~\citep{shimizu2021hyperbolic,van_Spengler_2023_ICCV},
\begin{equation}
\mathrm{PoincareFC}_{\mathbf{W},\mathbf{b}}(\mathbf{x})
\;=\; \exp_{\mathbf{o}}^{c}\!\bigl(\mathbf{W}\,\log_{\mathbf{o}}^{c}(\mathbf{x})\bigr) \oplus \mathbf{b}.
\label{eq:poincare-fc}
\end{equation}
This realises a linear transformation on the manifold via the tangent space at the origin, with standard parameterization for $\mathbf{W}$ and $\mathbf{b}$, allowing optimization.

\section{Methodology}
\label{sec:method}
\subsection{Architecture Overview}
\begin{figure*}[ht]
\begingroup\setlength{\abovecaptionskip}{2pt}\setlength{\belowcaptionskip}{1pt}
    \centering
    \includegraphics[width=\linewidth]{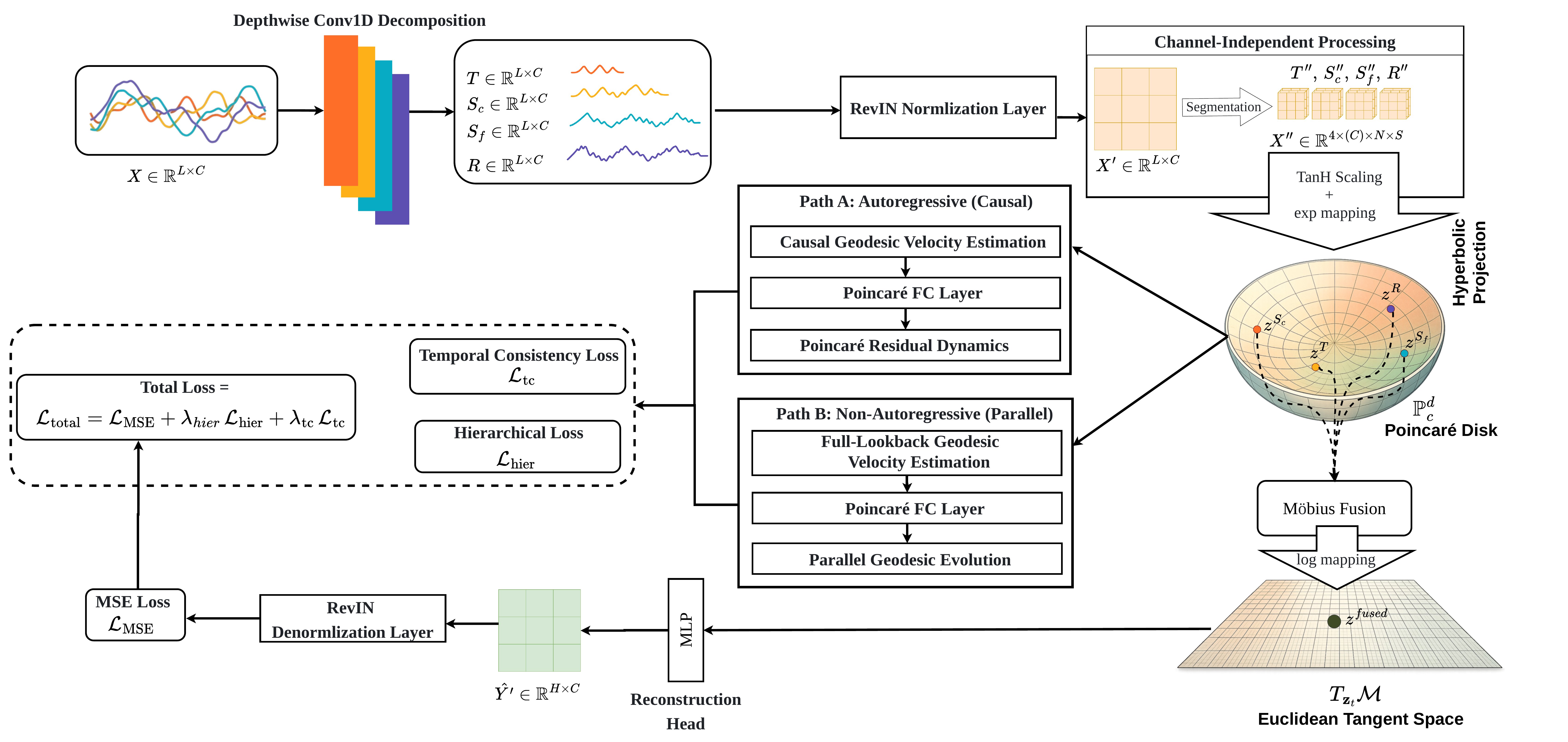}
    \caption{
    Overview of HyperForecast pipeline.
    Input $\mathbf{X}\!\in\!\mathbb{R}^{L\times C}$ is decomposed into trend $\mathbf{T}$, coarse seasonal $\mathbf{S}_c$, fine seasonal $\mathbf{S}_f$, and residual $\mathbf{R}$ components and normalized by a shared RevIN using statistics from the reconstructed signal $\mathbf{X}\!=\!\mathbf{T}\!+\!\mathbf{S}_c\!+\!\mathbf{S}_f\!+\!\mathbf{R}$. Each component is then segment-encoded, projected onto the Poincar\'e ball $\mathbb{P}^d_c$, forecast via geodesic dynamics on the manifold, and M\"obius-fused before Euclidean decoding.
    }
    \label{fig:hyperbolic-forecasting}
\endgroup
\end{figure*}

Given a multivariate time series $\mathbf{X} \in \mathbb{R}^{L \times C}$ with lookback length $L$ and $C$ channels (features), our goal is to forecast $\hat{\mathbf{Y}} \in \mathbb{R}^{H \times C}$ across next $H$ timesteps. HyperForecast performs forecasting in hyperbolic space through four stages: (1) learnable multi-scale decomposition of $\mathbf{X}$ into trend, coarse seasonal, fine seasonal, and residual components, (2) channel-independent segment encoding with projection onto the Poincar\'e ball $\mathbb{P}^d_c$, (3) forecasting via geodesic velocity either autoregressively (moving-window velocity estimation, sequential rollout) or non-autoregressively (full-lookback velocity, parallel prediction), over horizon $H_{\text{pred}} = \lfloor H / S \rfloor$ segments of fixed length $S$, and (4) M\"obius fusion of the components and Euclidean decoding. For clarity, we present the formulation for a single sample; all operations apply independently across batch samples during training. The complete pipeline is illustrated in Figure~\ref{fig:hyperbolic-forecasting}. 

\subsection{Learnable Decomposition}
\label{subsec:learnable_decomposition}

Given a multivariate time series $\mathbf{X} \in \mathbb{R}^{L \times C}$, we first decompose the signal using an additive decomposition inspired by MSTL~\citep{bandara2025mstl},
\begin{equation}
\mathbf{X} = \mathbf{T} + \mathbf{S}_c + \mathbf{S}_f + \mathbf{R},
\label{eq:decomp}
\end{equation}
where $\mathbf{T}$, $\mathbf{S}_c$,$\mathbf{S}_f$, and $\mathbf{R}$ denote the trend, coarse seasonal, fine seasonal, and residual components, respectively. Components are extracted via cascaded depthwise convolutions,
\begin{equation}
\begin{aligned}
\mathbf{T} &= \mathrm{Conv1D}_{\mathrm{trend}}(\mathbf{X}), &
\mathbf{S}_c &= \mathrm{Conv1D}_{\mathrm{coarse}}(\mathbf{X} - \mathbf{T}), \\
\mathbf{S}_f &= \mathrm{Conv1D}_{\mathrm{fine}}(\mathbf{X} - \mathbf{T} - \mathbf{S}_c), &
\mathbf{R} &= \mathbf{X} - \mathbf{T} - \mathbf{S}_c - \mathbf{S}_f.
\end{aligned}
\label{eq:cascade}
\end{equation}
This cascaded subtraction structure induces the ordering $\mathbf{T} \prec \mathbf{S}_c \prec \mathbf{S}_f \prec \mathbf{R}$, each component captures variation unexplained by the slower-varying components above it. Kernel sizes are initialized from the dataset sampling frequency with $k_c > k_f$, enabling coarse-to-fine modeling of periodic structures. A larger kernel averages over a wider temporal window, hence isolating slower-varying (coarser) periodicity, so that each successive component tends to capture higher-frequency content than the last, facilitating a nested structure that the radial loss subsequently encodes. All operations are differentiable and jointly optimized with the forecasting objective. Reversible instance normalization (RevIN) \citep{kim2022reversible} is applied using statistics computed from the reconstructed signal, ensuring statistical consistency and exact denormalization during reconstruction.

\subsection{Segment Encoding} 
Following normalization, each decomposed component $\mathbf{T}, \mathbf{S}_c, \mathbf{S}_f, \mathbf{R} \in \mathbb{R}^{L \times C}$ undergoes channel-independent processing. Channel-independent processing collapses the feature dimension into the batch axis, yielding $C$ univariate time series per component. Each component tensor is then partitioned along the temporal dimension into $N = {L}/{S}$ non-overlapping segments of fixed length $S$, i.e., $\mathbf{X} = \{\mathbf{X}_1, \mathbf{X}_2, \ldots,\mathbf{X}_N\}$.

We define a segment encoding operator $\Phi_S(\cdot)$ that maps each segment to a $d$-dimensional latent vector. Four separate linear encoders are used, one for each decomposed component, mapping segments to $d$ Euclidean latent vectors:
\begin{equation}
\mathbf{h}_{\mathbf{k}} = \Phi_S^{\mathbf{k}}(\mathbf{k}), \quad 
\mathbf{k} \in \{\mathbf{T}, \mathbf{S}_c, \mathbf{S}_f, \mathbf{R}\}.
\end{equation}
No parameters are shared across component encoders. Using separate encoders allows each component to learn representations tailored to its characteristic temporal scale and variability.

\paragraph{Projection to Hyperbolic Space.}
The Euclidean latent vectors are then mapped to the $d$-dimensional Poincar\'e
ball $\mathbb{P}^d_c$ using the exponential map at the origin: $ \mathbf{z} = \exp_{\mathbf{o}}\!\left(\gamma \cdot \mathbf{h}\right)$,
where $\gamma = \tanh(\theta)$ is a \textit{learnable scaling factor}, with 
$\theta$ initialized to $1.0$. The $\tanh$ activation constrains 
$\gamma \in (-1, 1)$ within the Poincar\'e ball radius, preventing numerical instability during the exponential map by bounding the tangent vector norm. This is important because the exponential map's closed form on the Poincar\'e ball involves hyperbolic tangent functions that saturate for large-norms. By adaptively scaling the Euclidean embeddings before projection, the model learns the appropriate ``depth'' in hyperbolic space for representing each component. This projection embeds segment-level representations into hyperbolic space, which is well-suited for modeling the hierarchical structure induced by multi-scale temporal decomposition. Once historical segments are encoded on $\mathbb{P}^d_c$, forecasting reduces to predicting the next $H_{\text{pred}} = \lfloor H/S \rfloor$ points along a manifold trajectory. We instantiate this in two variants that differ in their manifold step: an Autoregressive rollout (Sec.~\ref{sec:ar}) and a Non-Autoregressive parallel evolution (Sec.~\ref{sec:nar}). 


\subsection{Hyperbolic Autoregressive (AR) Forecasting}
\label{sec:ar}
Forecasting is performed independently for each component over horizon $H_{\text{pred}} = \lfloor H / S \rfloor$ corresponding to the number of future segments to be generated. At each prediction step $t$, a moving window of the previous $K$ hyperbolic embeddings is used to compute velocities via logarithmic maps between consecutive states. Unlike sliding windows over fixed input, the moving window dynamically incorporates newly predicted states during autoregressive generation to maintain temporal context for velocity estimation.

\paragraph{Causal Geodesic Velocity Estimation.}
At each step $t \in \{1, \ldots, H_{\text{pred}}\}$, we estimate a local geodesic velocity $\tilde{\mathbf{v}}_t$ using a causal moving window of the previous $K$ hyperbolic embeddings. Each displacement $\log_{\mathbf{z}_{t-i}}(\mathbf{z}_{t-i+1})$ lies in a different tangent space $T_{\mathbf{z}_{t-i}}\mathcal{M}$, so before aggregating we align every displacement into the shared tangent space $T_{\mathbf{z}_t}\mathcal{M}$ via parallel transport $\Gamma_{\mathbf{z}_{t-i}\to\mathbf{z}_t}$ (Eq.~\ref{eq:parallel-transport}). Specifically, $\tilde{\mathbf{v}}_t$ is defined as a weighted aggregation of parallel-transported logarithmic-map displacements between consecutive states within the window,
\begin{align}
\tilde{\mathbf{v}}_t = \sum_{i=0}^{K-2} w_i \, \Gamma_{\mathbf{z}_{t-1-i}\to\mathbf{z}_t}\!\left(\log_{\mathbf{z}_{t-1-i}}\!\left(\mathbf{z}_{t-i}\right)\right),
\label{eq:ar-vtilde}
\end{align}
where $\log_{\mathbf{z}_{t-1-i}}(\mathbf{z}_{t-i}) \in T_{\mathbf{z}_{t-1-i}}\mathcal{M}$ denotes the logarithmic map at base point $\mathbf{z}_{t-1-i}$, mapping the point $\mathbf{z}_{t-i}$ on the manifold to a tangent vector representing the geodesic displacement from $\mathbf{z}_{t-1-i}$ to $\mathbf{z}_{t-i}$, and $\Gamma_{\mathbf{z}_{t-1-i}\to\mathbf{z}_t}$ carries that displacement into the common reference frame $T_{\mathbf{z}_t}\mathcal{M}$ before summation. $w_i$ are normalized exponentially decaying weights that prioritize recent transitions. The resulting $\tilde{\mathbf{v}}_t$ provides a smoothed estimate of temporal evolution over a window of size $K$ on the manifold.
\paragraph{Poincar\'e Residual Dynamics.}
The aggregated velocity is transformed by a learnable linear map $\mathbf{W} \in \mathbb{R}^{d \times d}$ and scaled by a step-size parameter $\beta$ via sigmoid activation $\sigma$, yielding $\bar{\mathbf{v}}_t = \sigma(\beta) \cdot \mathbf{W} \tilde{\mathbf{v}}_t,$ where $\sigma(\beta) \in (0, 1)$ adaptively controls the extrapolation magnitude. The next hyperbolic state is predicted using Poincar\'e residual dynamics,
\begin{align}
\mathbf{z}_{t+1} = \bigl(\alpha \otimes \mathbf{z}_t\bigr) \oplus \bigl((1-\alpha) \otimes \exp_{\mathbf{z}_{t}}(\bar{\mathbf{v}}_t\bigr)),
\label{eq:ar-update}
\end{align}
where $\alpha \in (0, 1)$ is a learnable residual blending weight, $\otimes$ denotes M\"obius scalar multiplication, $\oplus$ denotes M\"obius addition, and $\exp_{\mathbf{z}_t}(\cdot)$ is the exponential map from $\mathbf{z}_t$. This formulation enables smooth extrapolation with this Poincar\'e residual connection to stabilize long-horizon forecasting along geodesic trajectories.

\subsection{Hyperbolic Non-Autoregressive (NAR) Forecasting}
\label{sec:nar}
The autoregressive branch in Section~\ref{sec:ar} predicts one segment per recurrent step. At every horizon $t$, it re-aggregates a window velocity for each component. It then pushes it through the Poincar\'e FC layer, and applies the Poincar\'e residual update. Although accurate, this approach is susceptible to error accumulation, since each step's prediction depends on the previous step, degrading performance at longer horizons. We therefore introduce a Non-Autoregressive (NAR) variant that produces all $H_{\text{pred}}$ future segments in parallel while reusing the same velocity mechanism.

\paragraph{Full-Lookback Geodesic Velocity Estimation.}

Let $(\mathbf{z}_1, \mathbf{z}_2, \dots, \mathbf{z}_N)$ denote the $N = L/S$ historical embeddings on $\mathbb{P}^d_c$ in temporal order, where $\mathbf{z}_N$ is the most-recent encoded historical segment. As in Eq.~\ref{eq:ar-vtilde}, each displacement is computed in the tangent space at its own base point, so we parallel-transport every term into the shared frame $T_{\mathbf{z}_N}\mathcal{M}$ before aggregating over the \emph{entire} lookback,

\begin{equation}
\tilde{\mathbf{v}}
\;=\;
\sum_{i=0}^{N-2} w_i \, \Gamma_{\mathbf{z}_{N-1-i}\to\mathbf{z}_N}\!\left(\log_{\mathbf{z}_{N-1-i}}\!\left(\mathbf{z}_{N-i}\right)\right),
\label{eq:nar-vtilde}
\end{equation}
which is structurally identical to Eq.~\ref{eq:ar-vtilde} but uses the full $N$-segment historical lookback instead of AR's $K$-segment moving window. We then pass $\tilde{\mathbf{v}}$ through the Poincar\'e FC layer (Eq.~\ref{eq:poincare-fc}) for each component. The two branches use the same velocity machinery (Eqs.~\ref{eq:ar-vtilde}-\ref{eq:poincare-fc}) but on different supports. AR re-aggregates over a moving window of $K$ segments at every rollout step ($K$), whereas NAR aggregates once over the entire $N = L/S$ historical encodings.

\paragraph{Parallel Geodesic Evolution.}
In place of the autoregressive Poincar\'e residual update of Eq.~\ref{eq:ar-update}, we extrapolate along the geodesic emitted from $\mathbf{z}_N$ with horizon-dependent step length, producing every future segment in a single batched call to $\exp_{\mathbf{z}_N}$,
\begin{align}
\mathbf{z}_{1:H_{\text{pred}}} \;=\; \exp_{\mathbf{z}_N}\!\bigl(t \cdot \bar{\mathbf{v}}\bigr),
\label{eq:nar-evolve}
\end{align}
where $t \in \{1, 2, \dots, H_{\text{pred}}\}$ and the $t$-th output is $\mathbf{z}_t = \exp_{\mathbf{z}_N}(t \cdot \bar{\mathbf{v}})$. 

\subsection{Hierarchical Constraints and Temporal Regularization}
To align learned representations with the frequency hierarchy, we employ two auxiliary losses during training.

\paragraph{Hierarchical Radial Loss.}
Component-wise radial constraints enforce frequency hierarchy by encouraging fine-scale and residual components to occupy larger radii in hyperbolic space while coarse-scale and trend components remain closer to the origin,
\begin{equation}
\mathcal{L}_{\text{radial}} = \sum_{k=1}^{3} \text{ReLU}\bigl(r_k - r_{k+1} + m\bigr),
\end{equation}
where $r_k = d(\mathbf{o}, \mathbf{z}^{(k)})$ denotes hyperbolic distance from the origin, with components ordered $k \in \{\mathbf{T}, \mathbf{S}_c, \mathbf{S}_f, \mathbf{R}\}$, and $m = 0.1$ is a fixed margin enforcing a minimum radial gap between consecutive components. A geodesic coherence term additionally enforces that adjacent components occupy neighboring manifold regions, reflecting the nested structure of the cascaded decomposition,
\begin{equation}
\mathcal{L}_{\text{hier}} = \mathcal{L}_{\text{radial}} + 0.5 \cdot \bigl[d(\mathbf{z}^\mathbf{T}, \mathbf{z}^{\mathbf{S}_c}) + d(\mathbf{z}^{\mathbf{S}_c}, \mathbf{z}^{\mathbf{S}_f}) + d(\mathbf{z}^{\mathbf{S}_f}, \mathbf{z}^\mathbf{R})\bigr].
\end{equation}

\paragraph{Temporal Consistency Loss.}
To promote smooth latent trajectories, we minimize abrupt velocity changes across prediction steps, promoting smooth and stable latent dynamics. Velocities are aligned via parallel transport $\Gamma$ before comparison,
\begin{equation}
\mathcal{L}_{\text{tc}}
=
\frac{1}{N-2}
\sum_{t=1}^{N-2}
\Bigl\|
\log_{\mathbf{z}_{t+1}}\!\left(\mathbf{z}_{t+2}\right)
-
\Gamma_{\mathbf{z}_t \to \mathbf{z}_{t+1}}
\!\left(
\log_{\mathbf{z}_t}\!\left(\mathbf{z}_{t+1}\right)
\right)
\Bigr\|^{2},
\label{eq:tc-loss}
\end{equation}
where $\Gamma_{\mathbf{z}_t \to \mathbf{z}_{t+1}}$ denotes parallel transport from tangent space $T_{\mathbf{z}_t}\mathcal{M}$ to $T_{\mathbf{z}_{t+1}}\mathcal{M}$. Trajectories with zero acceleration correspond to geodesics in hyperbolic space.

\paragraph{Hyperbolic Fusion and Reconstruction}
Predicted latent representations from all components are fused using M\"obius
addition, $\mathbf{z}^{\text{fused}} = \mathbf{z}^{\mathbf{T}} \oplus \mathbf{z}^{\mathbf{S}_c} \oplus \mathbf{z}^{\mathbf{S}_f} \oplus \mathbf{z}^{\mathbf{R}}$.
This integrates information from multiple temporal scales into a single hyperbolic latent representation. This fused representation is mapped to Euclidean space via logarithmic mapping and decoded through a single-layer MLP with ReLU activation~\citep{nair2010rectified} to produce the final forecast $\hat{\mathbf{Y}} \in \mathbb{R}^{H \times C}$. RevIN denormalization restores the original scale. The total training loss combines forecasting MSE with the auxiliary losses,
\begin{equation}
\mathcal{L} = \mathcal{L}_{\text{MSE}} + \lambda_{\text{hier}} \mathcal{L}_{\text{hier}} + \lambda_{\text{tc}} \mathcal{L}_{\text{tc}},
\end{equation}
where $\lambda_{\text{hier}}$ and $\lambda_{\text{tc}}$ are tuned via grid search per dataset.

\section{Experiments and Results}
\label{sec:experiments}
\paragraph{Datasets.}
We evaluate on 11 publicly available datasets spanning diverse domains, following standard LTSF evaluation practice with a fixed lookback length and multiple prediction horizons: \textbf{Weather}~\citep{lai2018modeling}, \textbf{Solar-Energy}~\citep{lai2018modeling}, \textbf{Exchange}~\citep{lai2018modeling}, \textbf{AQShunyi, AQWan}~\citep{zhang2017cautionary}, \textbf{CzeLan, ZafNoo}~\citep{poyatos2020global}, \textbf{PM2.5, Temp}~\citep{wang2020pm2}, \textbf{Wind}~\citep{li2022generative}, \textbf{METR-LA}~\citep{li2018diffusion}. Dataset statistics are summarized alongside the geometric diagnostics in Table~\ref{tab:datasets}. We use a sequence length $L=720$, and report results averaged over four standard horizons $H \in \{96,192,336,720\}$; implementation details, including hyperparameters, optimization settings, and computational requirements, are in Appendix~\ref{app:implementation_details}.

\paragraph{Results.}
We report MSE and MAE between $\mathbf{Y}$ and $\hat{\mathbf{Y}}\in\mathbb{R}^{H\times C}$, averaged over all $H\times C$ (lower is better). The averaged forecasting results are in Table~\ref{tab:main_results_mean}. HyperForecast NAR ranks first by MSE on 7 of 11 datasets and first-or-second on 10 of 11 (the exception is METR-LA, where PatchTST and iTransformer lead).
NAR outperforms AR on most datasets, with the gap most pronounced at long horizons on Temp (NAR: 0.113, AR: 0.203 at $H\!=\!720$), validating that parallel geodesic evolution avoids error accumulation in sequential rollout. Mean$\,\pm\,$std MSE over 5 seeds on seven representative datasets (Appendix~\ref{sec:stability}) confirms low variance across runs; a full efficiency comparison, including parameter/MACs breakdown and an AR-vs-NAR cost analysis, is in Appendix~\ref{sec:model_efficiency}, and full per-horizon results are in Appendix~\ref{sec:full_results}. We next examine the geometric structure of the learned embeddings underlying these gains.

\begin{table*}[h]
\begingroup\setlength{\abovecaptionskip}{2pt}\setlength{\belowcaptionskip}{1pt}
\centering
\scriptsize
\setlength{\tabcolsep}{2.2pt}
\renewcommand{\arraystretch}{1.0}
\caption{Average long-term forecasting results (MSE/MAE) across horizons $H\in\{96,192,336,720\}$ on 11 datasets (input length $L=720$ for all models). Best values are \textbf{\textcolor{red}{red-bold}}; second-best \textbf{\textcolor{blue}{blue-bold}}.}
\label{tab:main_results_mean}
\resizebox{\textwidth}{!}{%
\begin{tabular}{l | cc | cc | cc | cc | cc | cc | cc | cc | cc | cc | cc | cc | cc | cc}
\toprule
\textbf{Dataset} & \multicolumn{2}{c}{\textbf{HF NAR}} & 
\multicolumn{2}{c} {\textbf{HF AR}} &
\multicolumn{2}{c}{\textbf{PhaseFormer}} & \multicolumn{2}{c}{\textbf{TimeBase}} & \multicolumn{2}{c}{\textbf{SparseTSF}} & \multicolumn{2}{c}{\textbf{TimeMixer}} & \multicolumn{2}{c}{\textbf{FITS}} & \multicolumn{2}{c}{\textbf{iTransformer}} & \multicolumn{2}{c}{\textbf{DLinear}} & \multicolumn{2}{c}{\textbf{PatchTST}} & \multicolumn{2}{c}{\textbf{TimesNet}} & \multicolumn{2}{c}{\textbf{FEDformer}} & \multicolumn{2}{c}{\textbf{Autoformer}} & \multicolumn{2}{c}{\textbf{Informer}} \\
\cmidrule(lr){1-1}\cmidrule(lr){2-3}\cmidrule(lr){4-5}\cmidrule(lr){6-7}\cmidrule(lr){8-9}\cmidrule(lr){10-11}\cmidrule(lr){12-13}\cmidrule(lr){14-15}\cmidrule(lr){16-17}\cmidrule(lr){18-19}\cmidrule(lr){20-21}\cmidrule(lr){22-23}\cmidrule(lr){24-25}\cmidrule(lr){26-27}\cmidrule(lr){28-29}
 & MSE & MAE & MSE & MAE & MSE & MAE & MSE & MAE & MSE & MAE & MSE & MAE & MSE & MAE & MSE & MAE & MSE & MAE & MSE & MAE & MSE & MAE & MSE & MAE & MSE & MAE & MSE & MAE \\
\midrule

Weather 
& \textbf{\textcolor{red}{0.218}} & 0.273
& 0.221 & 0.270
& 0.223 & \textbf{\textcolor{red}{0.261}}
& \textbf{\textcolor{blue}{0.219}} & \textbf{\textcolor{blue}{0.263}}
& 0.244 & 0.285
& 0.238 & 0.281
& 0.244 & 0.286
& 0.233 & 0.273
& 0.242 & 0.299
& 0.223 & 0.264
& 0.254 & 0.293
& 0.355 & 0.391
& 0.446 & 0.457
& 0.567 & 0.513 \\

AQShunyi 
& \textbf{\textcolor{red}{0.657}} & \textbf{\textcolor{red}{0.503}}
& \textbf{\textcolor{blue}{0.665}} & \textbf{\textcolor{blue}{0.504}}
& 0.722 & 0.523
& 0.674 & 0.506
& 0.760 & 0.546
& 0.736 & 0.536
& 0.763 & 0.548
& 0.722 & 0.520
& 0.757 & 0.572
& 0.690 & \textbf{\textcolor{red}{0.503}}
& 0.788 & 0.559
& 1.130 & 0.757
& 1.393 & 0.878
& 1.688 & 0.966 \\

AQWan 
& \textbf{\textcolor{red}{0.751}} & \textbf{\textcolor{red}{0.486}}
& \textbf{\textcolor{blue}{0.762}} & \textbf{\textcolor{blue}{0.495}}
& 0.837 & 0.517
& 0.779 & 0.499
& 0.826 & 0.526
& 0.822 & 0.521
& 0.813 & 0.519
& 0.843 & 0.538
& 0.883 & 0.560
& 0.810 & 0.513
& 0.857 & 0.543
& 0.931 & 0.565
& 1.001 & 0.654
& 1.123 & 0.744 \\

CzeLan 
& \textbf{\textcolor{red}{0.221}} & 0.286
& \textbf{\textcolor{blue}{0.224}} & 0.286
& 0.225 & \textbf{\textcolor{blue}{0.273}}
& 0.225 & \textbf{\textcolor{red}{0.262}}
& 0.240 & 0.292
& 0.234 & 0.282
& 0.250 & 0.304
& 0.245 & 0.296
& 0.241 & 0.290
& 0.229 & 0.275
& 0.258 & 0.312
& 0.266 & 0.314
& 0.289 & 0.368
& 0.347 & 0.426 \\

ZafNoo 
& \textbf{\textcolor{red}{0.485}} & \textbf{\textcolor{red}{0.443}}
& \textbf{\textcolor{blue}{0.489}} & 0.448
& 0.533 & 0.456
& 0.495 & \textbf{\textcolor{blue}{0.445}}
& 0.531 & 0.490
& 0.505 & 0.464
& 0.520 & 0.480
& 0.540 & 0.500
& 0.540 & 0.495
& 0.510 & 0.469
& 0.545 & 0.502
& 0.601 & 0.539
& 0.667 & 0.635
& 0.726 & 0.697 \\

Exchange 
& \textbf{\textcolor{blue}{0.289}} & 0.408
& \textbf{\textcolor{red}{0.274}} & \textbf{\textcolor{red}{0.379}}
& 0.504 & 0.452
& 0.309 & \textbf{\textcolor{blue}{0.392}}
& 0.316 & 0.410
& 0.330 & 0.426
& 0.302 & 0.400
& 0.301 & 0.401
& 0.311 & 0.410
& 0.308 & 0.400
& 0.327 & 0.425
& 0.380 & 0.472
& 0.447 & 0.589
& 0.432 & 0.581 \\

METR-LA 
& 1.266 & 0.741
& 1.296 & 0.754
& 1.348 & 0.762
& 1.254 & 0.754
& 1.302 & 0.764
& 1.270 & 0.741
& 1.296 & 0.760
& \textbf{\textcolor{blue}{1.253}} & \textbf{\textcolor{blue}{0.734}}
& 1.300 & 0.759
& \textbf{\textcolor{red}{1.210}} & \textbf{\textcolor{red}{0.706}}
& 1.390 & 0.812
& 1.550 & 0.875
& 1.639 & 0.974
& 1.893 & 1.146 \\
 
PM2.5 
& \textbf{\textcolor{blue}{0.410}} & 0.452
& \textbf{\textcolor{red}{0.406}} & \textbf{\textcolor{blue}{0.437}}
& 0.430 & \textbf{\textcolor{red}{0.434}}
& 0.428 & 0.437
& 0.467 & 0.477
& 0.465 & 0.474
& 0.460 & 0.469
& 0.480 & 0.491
& 0.449 & 0.457
& 0.459 & 0.466
& 0.466 & 0.475
& 0.559 & 0.547
& 0.621 & 0.650
& 0.636 & 0.663 \\
 
Solar 
& \textbf{\textcolor{blue}{0.205}} & 0.274
& 0.211 & 0.268
& \textbf{\textcolor{red}{0.193}} & \textbf{\textcolor{red}{0.240}}
& 0.215 & \textbf{\textcolor{blue}{0.254}}
& 0.216 & 0.264
& 0.244 & 0.296
& 0.229 & 0.279
& 0.233 & 0.285
& 0.227 & 0.276
& 0.227 & 0.275
& 0.243 & 0.295
& 0.252 & 0.291
& 0.306 & 0.381
& 0.327 & 0.411 \\

Temp 
& \textbf{\textcolor{red}{0.119}} & \textbf{\textcolor{red}{0.267}}
& \textbf{\textcolor{blue}{0.145}} & \textbf{\textcolor{blue}{0.293}}
& 0.217 & 0.348
& 0.187 & 0.338
& 0.295 & 0.412
& 0.302 & 0.423
& 0.315 & 0.442
& 0.302 & 0.423
& 0.292 & 0.418
& 0.291 & 0.406
& 0.303 & 0.427
& 0.343 & 0.456
& 0.404 & 0.587
& 0.471 & 0.682 \\

Wind 
& \textbf{\textcolor{red}{0.930}} & \textbf{\textcolor{red}{0.700}}
& 0.949 & 0.713
& 1.018 & 0.743
& \textbf{\textcolor{blue}{0.940}} & 0.709
& 1.014 & 0.736
& 1.044 & 0.758
& 1.023 & 0.745
& 1.025 & 0.748
& 1.018 & 0.739
& 0.968 & \textbf{\textcolor{blue}{0.703}}
& 1.101 & 0.799
& 1.230 & 0.841
& 1.237 & 0.947
& 1.381 & 1.033 \\

\bottomrule
\end{tabular}
}
\endgroup
\end{table*}

\begin{figure*}[h]
\begingroup\setlength{\abovecaptionskip}{2pt}\setlength{\belowcaptionskip}{1pt}
\centering
\includegraphics[width=0.8\textwidth]{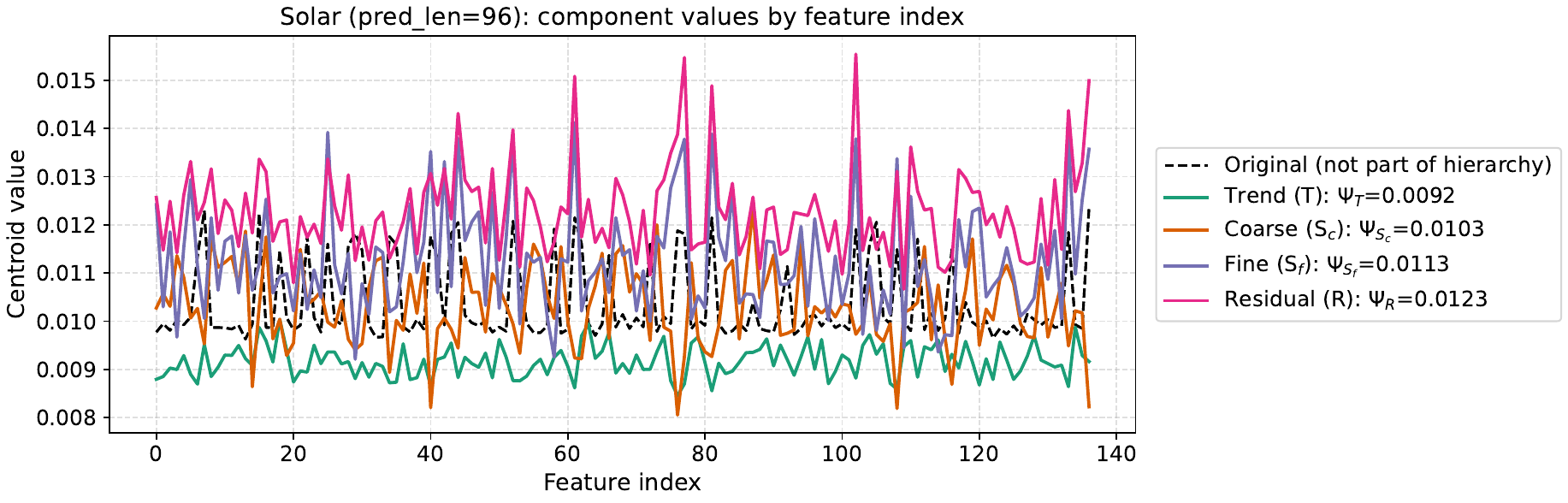}
\caption{Per-feature Welch-PSD spectral centroid of each decomposed component on Solar (137 channels, $H=96$). Black dashed: original (undecomposed) signal. Legend values are the per-component average across all 137 features. The ordering $T\!<\!S_c\!<\!S_f\!<\!R$ holds on the dataset average and for the large majority of individual features.}
\label{fig:centroid-solar}
\endgroup
\end{figure*}

\paragraph{Geometric Evidence for the Frequency Hierarchy.}
\label{sec:radial_hierarchy}
To understand why HyperForecast's forecasting gains concentrate on certain datasets, we examine three complementary diagnostics of the learned frequency hierarchy: the Welch-PSD~\citep{welch1967use} spectral centroid $\Psi$ of each decomposed component, the paired norm ratio $\rho$ (whether embedding radii preserve the trend$\prec$coarse$\prec$fine$\prec$residual ordering), and Gromov $\delta$-hyperbolicity $\delta_D$ (overall tree-likeness of the embedding geometry). Formal definitions of $\Psi$, $\rho$, and $\delta_D$ are given in Appendix~\ref{app:metrics-defs}. We first demonstrate whether the raw decomposed signal itself already exhibits the target frequency ordering, independent of any learned embedding (Figure~\ref{fig:centroid-solar} and Appendix~\ref{app:centroid-figures}); we then examine whether the trained hyperbolic embeddings inherit this ordering radially ($\rho$) and whether the resulting geometry is broadly tree-like ($\delta_D$), both reported per dataset alongside basic dataset statistics in Table~\ref{tab:datasets}.

As per our illustration of decomposition on Solar at $H=96$ in Figure~\ref{fig:centroid-solar}, average centroids $\Psi_T\!=\!0.0092 < \Psi_{S_c}\!=\!0.0103 < \Psi_{S_f}\!=\!0.0113 < \Psi_R\!=\!0.0123$, with the full ordering holding for 69\% of individual features. We also illustrate this trend for additional datasets spanning channel counts from 11 to 207 and different domains in Appendix~\ref{app:centroid-figures}, all exhibiting the full ordering on the dataset average. The cascaded decomposition induces a genuine frequency separation on the raw signal, prior to and independent of any hyperbolic embedding, establishing the geometric precondition that HyperForecast's radial design is built to leverage.

\begin{table*}[h]
\centering
\scriptsize
\caption{Dataset statistics and frequency-hierarchy geometric diagnostics for the 11 LTSF benchmarks. \textbf{Var}: number of variables. \textbf{Length}: total series length. \textbf{Freq}: sampling frequency. $\rho$: paired norm ratio between consecutive components' embedding radii ($\rho\!<\!1$ indicates good hierarchical separation). $\delta_D$: Gromov $\delta$-hyperbolicity of the trained embeddings, both averaged across $H\in\{96,192,336,720\}$.}
\label{tab:datasets}
\renewcommand{\arraystretch}{0.9}
\setlength{\tabcolsep}{4.2pt}
\begin{tabular}{lccccc@{\hspace{8pt}}|@{\hspace{8pt}}lccccc}
\toprule
Dataset & Var & Length & Freq & $\rho$ & $\delta_D$ & Dataset & Var & Length & Freq & $\rho$ & $\delta_D$ \\
\midrule
Weather  & 21 & 52,696 & 10m & 0.737 & 0.0139 & METR-LA & 207 & 34,272 & 5m  & 1.023 & 0.0143 \\
AQShunyi & 11 & 35,064 & 1h  & 1.127 & 0.0095 & PM2.5   & 184 & 11,688 & 3h  & 0.997 & 0.0194 \\
AQWan    & 11 & 35,064 & 1h  & 0.848 & 0.0200 & Solar   & 137 & 52,560 & 10m & 0.897 & 0.0150 \\
CzeLan   & 11 & 19,934 & 30m & 0.749 & 0.0025 & Temp    & 184 & 11,688 & 3h  & 1.020 & 0.0227 \\
ZafNoo   & 11 & 19,225 & 30m & 0.515 & 0.0021 & Wind    & 7   & 48,673 & 15m & 0.935 & 0.0153 \\
Exchange & 8  & 7,588  & 1d  & 0.886 & 0.0097 &         &     &        &     &       &       \\
\bottomrule
\end{tabular}
\renewcommand{\arraystretch}{1}
\end{table*}

The \textbf{paired norm ratio} $\rho$ (Table~\ref{tab:datasets}) falls below 1 in 8 of 11 datasets, i.e., the trained embeddings preserve the specific component ordering (trend nearest the origin, residual farthest) in most cases. \textbf{Gromov $\delta$-hyperbolicity} $\delta_D$ is close to 0 for all 11 datasets (range 0.002--0.023), consistent with an overall tree-like embedding geometry favoring hyperbolic space; since $\delta_D$ pools all four components into a single point cloud, it alone does not test whether the ordering is preserved, which is what $\rho$ captures. Elevated $\rho$ does not uniformly translate into weaker forecasting, AQShunyi ($\rho=1.127$) and Temp ($\rho=1.020$) still rank first by MSE (Table~\ref{tab:main_results_mean}). METR-LA is the only dataset where  HyperForecast does not rank first or second, plausibly due to its high channel count (207) and length in addition to its radial geometry ($\rho=1.023$).

\begin{figure*}[!h]
\begingroup\setlength{\abovecaptionskip}{2pt}\setlength{\belowcaptionskip}{1pt}
\centering
\includegraphics[width=\textwidth]{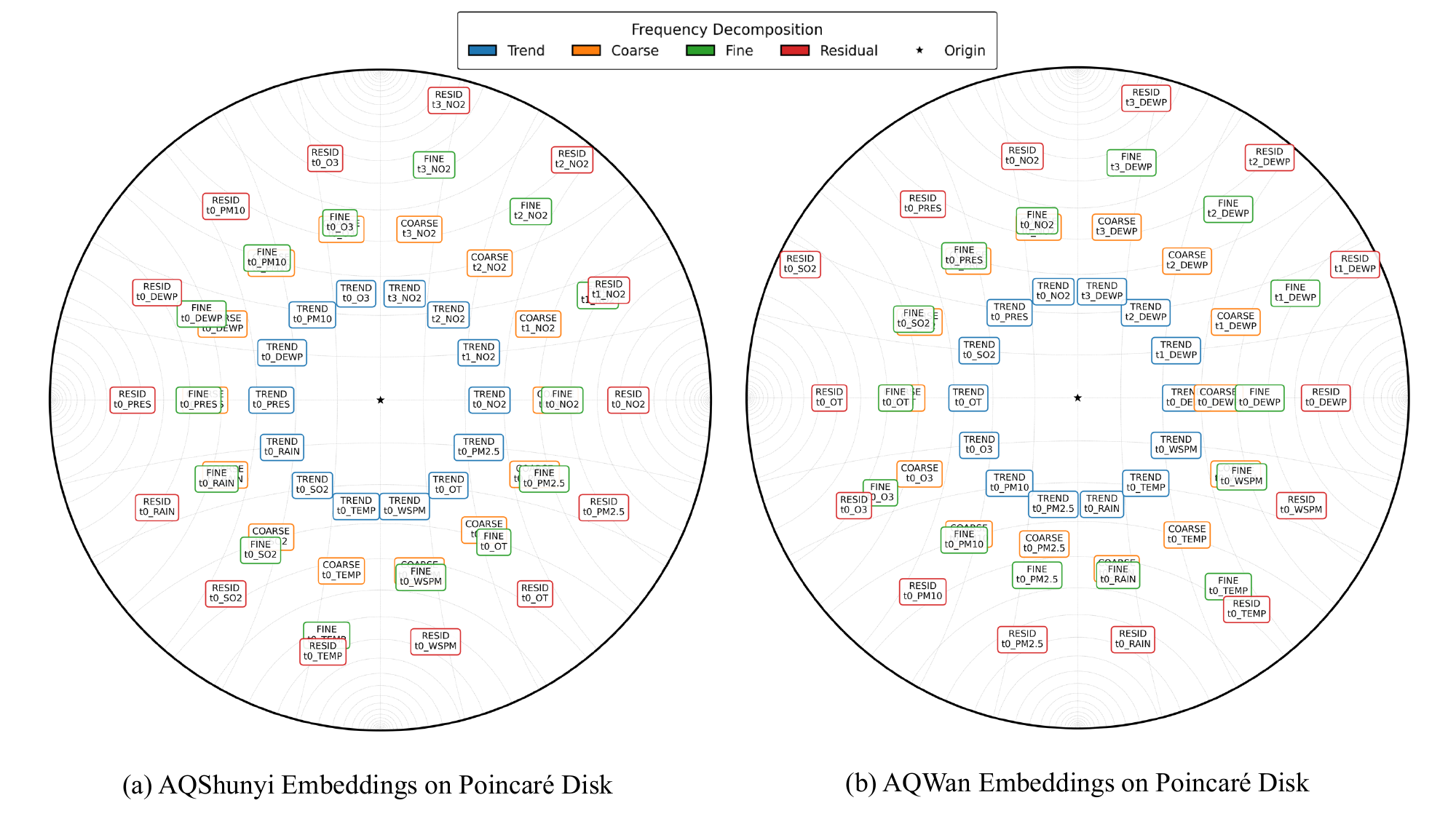}
\caption{Each point is a (Scale, TimeSegment, Feature) tuple. The \textbf{radial hierarchy} shows our model learns frequency-based decomposition (Trend $<$ Coarse $<$ Fine $<$ Residual). Labels follow \texttt{t\{segment\}\_\{feature\}} (e.g., t0\_Feature\_1 = segment 0, feature 1).}
\label{fig:plot2}
\endgroup
\end{figure*}

To validate the consistency of learned frequency-based hierarchies across diverse datasets, we visualize embeddings for AQShunyi and AQWan on the Poincar\'e disk (Figure~\ref{fig:plot2}). Both datasets exhibits moderate radial separation, with trend components clustering near the origin and residual components distributed toward the boundary. The overall structure maintains a tree-like radial organization consistent with frequency hierarchy, although the strict ordering may not hold on average for all samples (AQShunyi's $\rho=1.127$). 
To confirm this ordering is a geometric property of embedding rather than an effect of the radial loss, Appendix~\ref{app:hierloss-ablation} reports the visualization with $\lambda_{\text{hier}}=0$.

\paragraph{Ablations.}
We present two targeted ablations validating core design choices: the benefit of hyperbolic over Euclidean geometry, and sensitivity to lookback window length. Additional ablations (impact of learnable decomposition, channel-independent vs.\ channel-dependent processing, window size, loss weights) are provided in Appendix~\ref{sec:ablations}.

\begin{figure*}[h]
\begingroup\setlength{\abovecaptionskip}{2pt}\setlength{\belowcaptionskip}{1pt}
    \centering
    \begin{subfigure}[h]{\textwidth}
        \centering
        \includegraphics[width=0.8\textwidth]{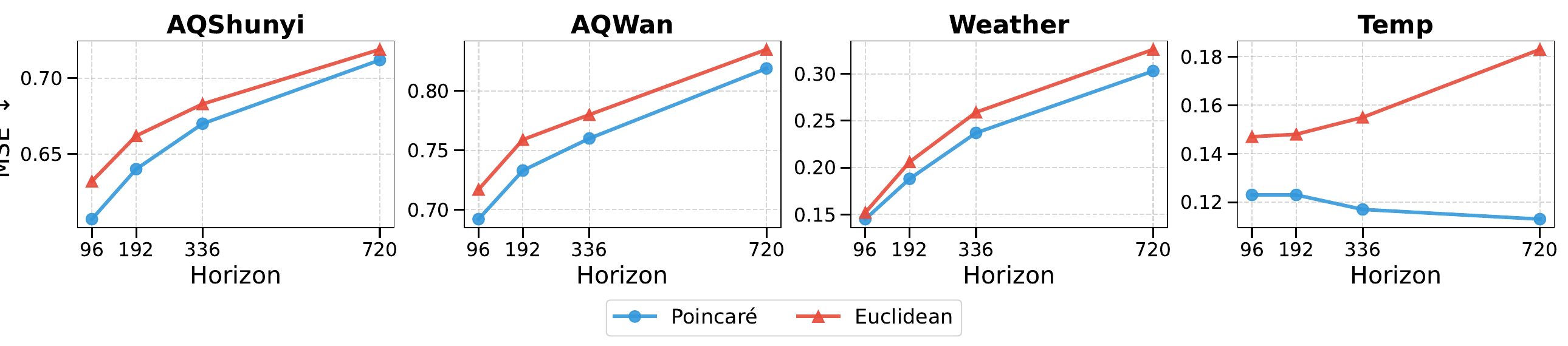}
        \caption{Poincar\'e vs.\ Euclidean geometry on AQShunyi, AQWan, Weather, and Temp (NAR).}
        \label{fig:ablation_poincare}
    \end{subfigure}

    \begin{subfigure}[h]{\textwidth}
        \centering
        \includegraphics[width=0.8\textwidth]{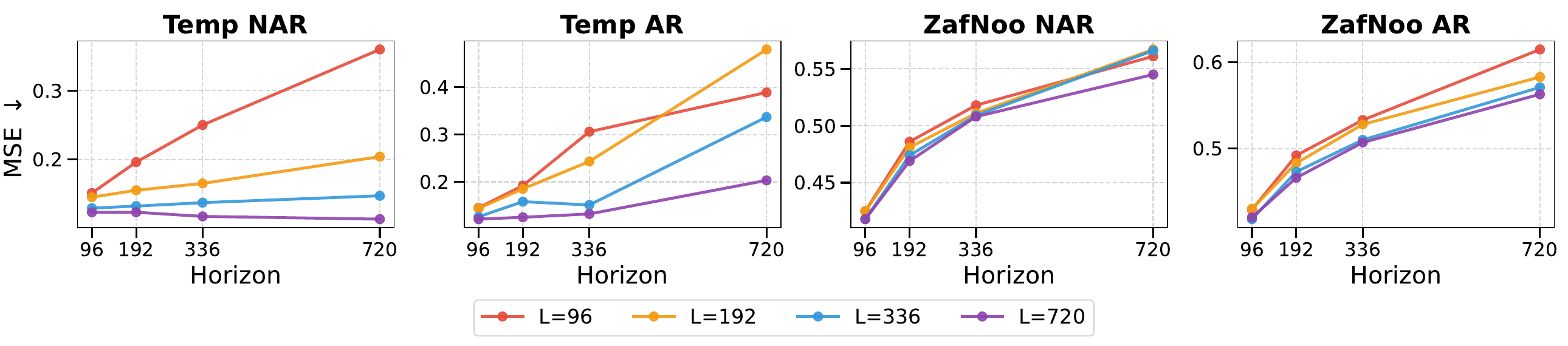}
        \caption{Lookback sensitivity ($L \in \{96,192,336,720\}$) for Temp and ZafNoo, NAR and AR.}
        \label{fig:ablation_lookback}
    \end{subfigure}

    \caption{Ablation studies: (a) Hyperbolic vs.\ Euclidean geometry and (b) lookback window sensitivity, shown for both forecasting variants.}
    \label{fig:ablation}
\endgroup
\end{figure*}

\ablationsub{Poincar\'e vs.\ Euclidean Geometry:}
We replace all manifold operations with standard Euclidean arithmetic, retaining identical decomposition and velocity dynamics. As shown in Figure~\ref{fig:ablation_poincare}, Poincar\'e embeddings achieve consistently lower MSE than their Euclidean counterpart across all horizons on all four datasets tested. 
Temp shows the largest and most horizon-dependent gap, where the Poincar\'e variant's MSE actually \emph{improves} with horizon (0.123$\to$0.113) while the Euclidean variant steadily degrades (0.147$\to$0.183), a 62\% higher MSE at $H=720$. 

\ablationsub{Lookback Window Sensitivity:}
The effect of input context length $L \in \{96, 192, 336, 720\}$ on NAR and AR variants across Temp and ZafNoo can be seen in Figure~\ref{fig:ablation_lookback}. On Temp, longer lookback provides a decisive advantage, with $L\!=\!720$ remaining consistently better while $L\!=\!96$ degrades sharply at longer horizons, confirming that richer context improves geodesic velocity estimation. 

\section{Conclusion}
\label{sec:conclusion}
This paper introduced HyperForecast, the first long-term time series forecasting framework operating in hyperbolic space. By treating decomposed segments as manifold trajectories that encode the frequency hierarchy the decomposition induces, HyperForecast achieves state-of-the-art performance on 9 of 11 benchmarks, with up to 58\% MSE reduction (Temp, $H{=}720$) over the strongest Euclidean baseline, and yields interpretable hyperbolic radii that confirm the learned decomposition.
Limitations include higher computational cost than ultra-lightweight models and the additive decomposition assumption, which may not hold for all time series. Future work could explore learnable curvature adaptation, hyperbolic probabilistic forecasting, and theoretical analysis of hyperbolic capacity for frequency hierarchies.

\section*{Acknowledgments.}
This work was supported by the DOE SEA-CROGS project (DE-SC0023191) and AFOSR project (FA9550-24-1-0231). We also thank the computing resources provided by the High Performance Computing (HPC) facility at NJIT.

\bibliographystyle{unsrtnat}
\bibliography{ref}

\newpage
\appendix
\section{Hyperbolic Geometry: Additional Definitions}
\label{app:hyperbolic-defs}

This appendix complements Section~\ref{sec:preliminaries} with the remaining hyperbolic operations referenced in the methodology. All formulations are stated for the Poincar\'e ball $\mathbb{P}^d_c$ of curvature $-c<0$, equipped with the conformal factor $\lambda_{\mathbf{x}} = \tfrac{2}{1 - c\lVert\mathbf{x}\rVert^2}$ from Sec.~\ref{sec:preliminaries}.

\paragraph{M\"obius addition and scalar multiplication.}
To perform neural network operations on $\mathbb{P}^d_c$, we use M\"obius addition and scalar multiplication as foundational gyrovector arithmetic~\citep{ungar2022gyrovector, ganea2018hyperbolic}:
\begin{align}
    \mathbf{x} \oplus \mathbf{y} 
        &= \frac{\left(1 + 2c\langle \mathbf{x}, \mathbf{y} \rangle 
                + c\lVert \mathbf{y} \rVert^2\right)\mathbf{x} 
                + \left(1 - c\lVert \mathbf{x} \rVert^2\right)\mathbf{y}}
               {1 + 2c\langle \mathbf{x}, \mathbf{y} \rangle 
                + c^2\lVert \mathbf{x} \rVert^2 \lVert \mathbf{y} \rVert^2},
        \label{eq:mobius_add} \\[6pt]
    \alpha \otimes_c \mathbf{x} 
        &= \tanh\!\left(\alpha\,\tanh^{-1}\!\left(
           \sqrt{c}\,\lVert \mathbf{x} \rVert\right)\right)
           \frac{\mathbf{x}}{\sqrt{c}\,\lVert \mathbf{x} \rVert + \varepsilon}.
        \label{eq:mobius_scalar}
\end{align}

\paragraph{Exponential and logarithmic maps at an arbitrary base point.}
For $\mathbf{x} \in \mathcal{M}$ and $\mathbf{v} \in T_{\mathbf{x}}\mathcal{M}$, the exponential map $\exp_{\mathbf{x}}^{c}\!:T_{\mathbf{x}}\mathcal{M} \!\to\! \mathcal{M}$ and its inverse $\log_{\mathbf{x}}^{c}\!:\mathcal{M} \!\to\! T_{\mathbf{x}}\mathcal{M}$ admit the closed-form expressions~\citep{ganea2018hyperbolic},
\begin{align}
\exp_{\mathbf{x}}^{c}(\mathbf{v})
&= \mathbf{x} \oplus \!\left(\tanh\!\bigl(\tfrac{\sqrt{c}\,\lambda_{\mathbf{x}}\,\lVert\mathbf{v}\rVert}{2}\bigr)\,
   \frac{\mathbf{v}}{\sqrt{c}\,\lVert\mathbf{v}\rVert + \varepsilon}\right),
\label{eq:exp-x}\\
\log_{\mathbf{x}}^{c}(\mathbf{y})
&= \frac{2}{\sqrt{c}\,\lambda_{\mathbf{x}}}\,
   \tanh^{-1}\!\bigl(\sqrt{c}\,\lVert -\mathbf{x}\oplus\mathbf{y}\rVert\bigr)\,
   \frac{-\mathbf{x}\oplus\mathbf{y}}{\lVert -\mathbf{x}\oplus\mathbf{y}\rVert + \varepsilon},
\label{eq:log-x}
\end{align}
where $\oplus$ is M\"obius addition (Eq.~\ref{eq:mobius_add}). Setting $\mathbf{x}=\mathbf{o}$ recovers the origin-centered formulas already given in Eq.~\ref{eq:log_map}. Exponential mapping is invoked at every AR rollout step (Eq.~\ref{eq:ar-update}) and once per horizon in the NAR branch (Eq.~\ref{eq:nar-evolve}), whereas logarithmic mapping is used in the velocity aggregators.

\paragraph{Hyperbolic distance.}
The geodesic distance between two arbitrary points $\mathbf{x}, \mathbf{y} \in \mathcal{M}$ is,
\begin{equation}
d_c(\mathbf{x}, \mathbf{y}) \;=\; \frac{2}{\sqrt{c}}\,\tanh^{-1}\!\bigl(\sqrt{c}\,\lVert -\mathbf{x}\oplus\mathbf{y}\rVert\bigr).
\label{eq:hyp-dist}
\end{equation}
At $\mathbf{x}=\mathbf{o}$, $d_c(\mathbf{o},\mathbf{y}) = \tfrac{2}{\sqrt{c}}\tanh^{-1}(\sqrt{c}\lVert\mathbf{y}\rVert)$ recovers the radial position.

\paragraph{Parallel transport.}
The parallel transport $\Gamma_{\mathbf{x}\to\mathbf{y}}\!:T_{\mathbf{x}}\mathcal{M} \!\to\! T_{\mathbf{y}}\mathcal{M}$ moves a tangent vector along the unique geodesic from $\mathbf{x}$ to $\mathbf{y}$ while preserving the Riemannian inner product. On $\mathcal{M}$ it has the closed form~\citep{ganea2018hyperbolic},
\begin{equation}
\Gamma_{\mathbf{x}\to\mathbf{y}}(\mathbf{v}) \;=\; \frac{\lambda_{\mathbf{x}}}{\lambda_{\mathbf{y}}}\,\mathrm{gyr}\!\bigl[\mathbf{y},\,-\mathbf{x}\bigr]\,\mathbf{v},
\label{eq:parallel-transport}
\end{equation}
where $\mathrm{gyr}[\mathbf{y},-\mathbf{x}]$ is the gyration that rotates the tangent vector along the geodesic from $\mathbf{x}$ to $\mathbf{y}$~\citep{ungar2022gyrovector}. Eq.~\ref{eq:parallel-transport} is invoked in the temporal-consistency loss $\mathcal{L}_{\mathrm{tc}}$ (Eq.~\ref{eq:tc-loss}) to align velocities sampled at different base points along the predicted trajectory before computing the acceleration penalty.

\section{Frequency-Hierarchy Metrics: Formal Definitions}
\label{app:metrics-defs}
This section formally defines the three per-dataset diagnostics discussed in Sec.~\ref{sec:radial_hierarchy} that are only described informally in the main text. The spectral centroid $\Psi$ is reported among illustrations in Figure~\ref{fig:centroid-solar} and Appendix~\ref{app:centroid-figures}. The paired norm ratio $\rho$, and Gromov $\delta$-hyperbolicity $\delta_D$ (both reported per dataset in Table~\ref{tab:datasets}).
 
\paragraph{Spectral centroid.} For a decomposed component $k \in \{\mathbf{T}, \mathbf{S}_c, \mathbf{S}_f, \mathbf{R}\}$, let $P_k(f)$ denote its Welch power spectral density~\citep{welch1967use}, estimated on the raw time-domain component (prior to any hyperbolic embedding), averaged across all samples, and computed per channel. The spectral centroid is the power-weighted mean frequency,
\begin{equation}
\Psi_k = \frac{\sum_f f \cdot P_k(f)}{\sum_f P_k(f)},
\label{eq:spectral-centroid}
\end{equation}
where higher $\Psi_k$ indicates higher-frequency content concentrated in component $k$.
 
\paragraph{Paired norm ratio.} Let $r_k(s) = d(\mathbf{o}, \mathbf{z}^{(k)}_s)$ denote the hyperbolic distance (Eq.~\ref{eq:hyp-dist}) of the $s$-th segment's trained embedding for component $k$ from the origin $\mathbf{o}$. For each of the three consecutive component pairs $(k,k') \in \{(\mathbf{T},\mathbf{S}_c), (\mathbf{S}_c,\mathbf{S}_f), (\mathbf{S}_f,\mathbf{R})\}$, we compute the per-segment ratio $r_k(s)/r_{k'}(s)$ and take its median over segments (median rather than mean, since segments embedded very close to the origin produce unstable individual ratios). The paired norm ratio $\rho$ is the mean of these three step-wise medians,
\begin{equation}
\rho = \frac{1}{3}\sum_{(k,k')} \operatorname{median}_s \left[\frac{r_k(s)}{r_{k'}(s)}\right],
\label{eq:norm-ratio}
\end{equation}
$\rho < 1$ indicates each successive, higher-frequency component sits farther from the origin than its predecessor, consistent with hierarchical radial separation. $\rho \geq 1$ indicates the ordering is not preserved on average.
 
\paragraph{Gromov $\delta$-hyperbolicity.} We measure each dataset's Gromov $\delta$-hyperbolicity~\citep{gromov1987hyperbolic} on the trained historical encoder embeddings, pooling all four decomposed components into a single set of points (unlike $\rho$ above, this diagnostic does not use component identity, so it tests overall tree-likeness rather than the specific component ordering). For a metric space $(X, d)$ -- here $d$ is hyperbolic distance (Eq.~\ref{eq:hyp-dist}) between pooled segment embeddings -- and any quadruple $x, y, z, w \in X$, let $S_{(1)} \ge S_{(2)} \ge S_{(3)}$ be the sorted pair-sums of $\{d(x,y){+}d(z,w),\, d(x,z){+}d(y,w),\, d(x,w){+}d(y,z)\}$. We define
\begin{equation}
\delta(X) = \tfrac{1}{2}\!\!\sup_{x,y,z,w \in X}\!\!\bigl(S_{(1)} - S_{(2)}\bigr),
\qquad
\delta_{D} = \delta(X)/D,
\label{eq:gromov-delta}
\end{equation}
where $D$ is the diameter of $X$. We normalize $\delta$ by $D$ following the relative-hyperbolicity convention~\citep{adcock2013tree,jonckheere2008scaled}: $\delta_D \!\to\! 0$ indicates a tree-like structure that strongly benefits from hyperbolic geometry, while $\delta_D \!\to\! 0.5$ indicates a flat/Euclidean structure.

\section{Model Efficiency Analysis}
\label{sec:model_efficiency}

We benchmark HyperForecast against all baselines under a single, unified protocol. All fourteen models here are run on Weather with $\text{seq\_len} = \text{pred\_len} = 720$. The parameter count, MACs, and FLOPs are measured for a single forward pass over one sample. The peak memory (Max Mem.) is measured at batch size 12, and epoch time and CPU inference time are wall-clock measurements under this same protocol.

\begin{figure}[h]
    \centering
    \begin{subfigure}[h]{0.68\linewidth}
        \centering
        \includegraphics[width=\linewidth]{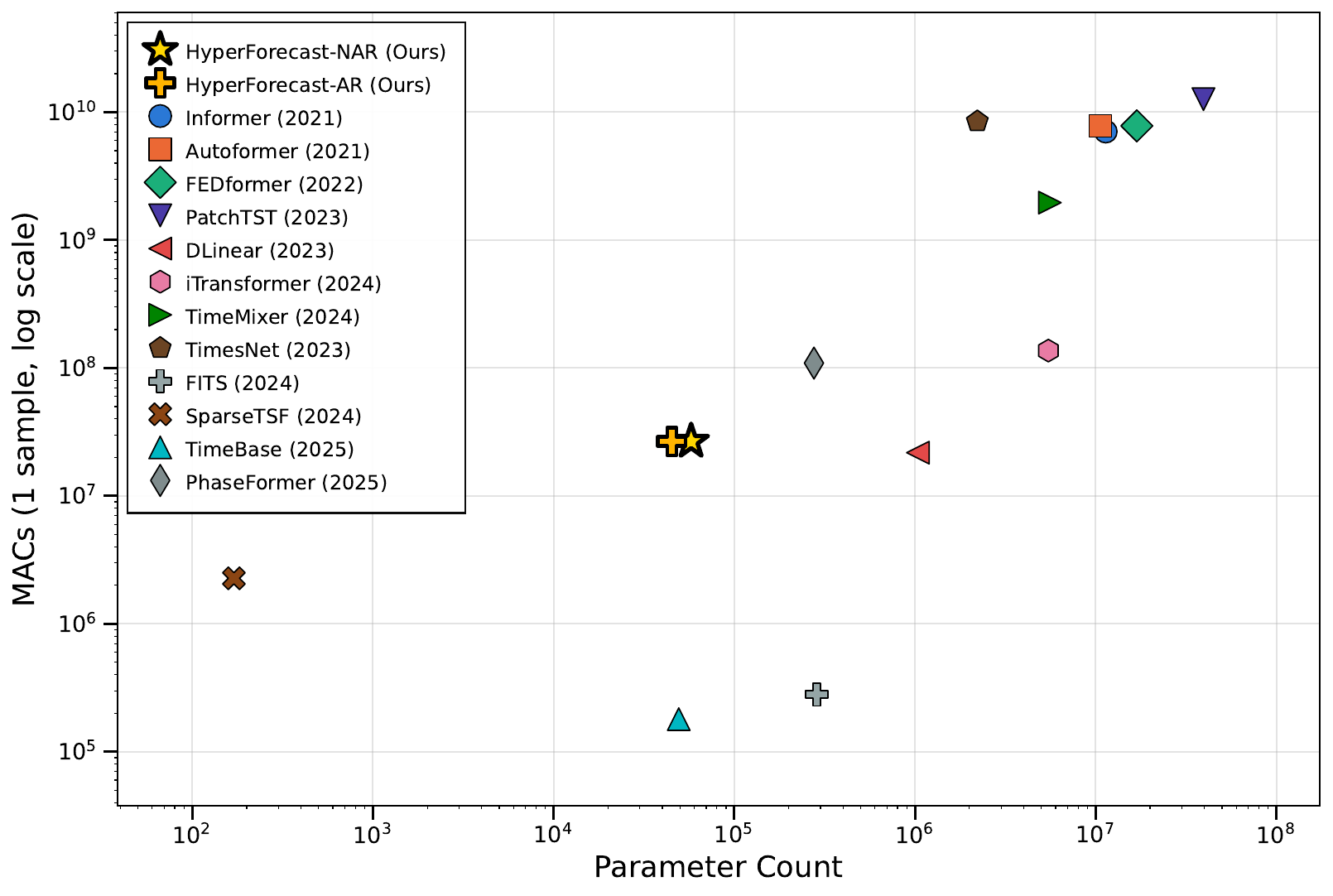}
        \caption{Parameter count vs.\ MACs (log--log).}
        \label{fig:efficiency}
    \end{subfigure}
    \begin{subfigure}[h]{0.83\linewidth}
        \centering
        \includegraphics[width=\linewidth]{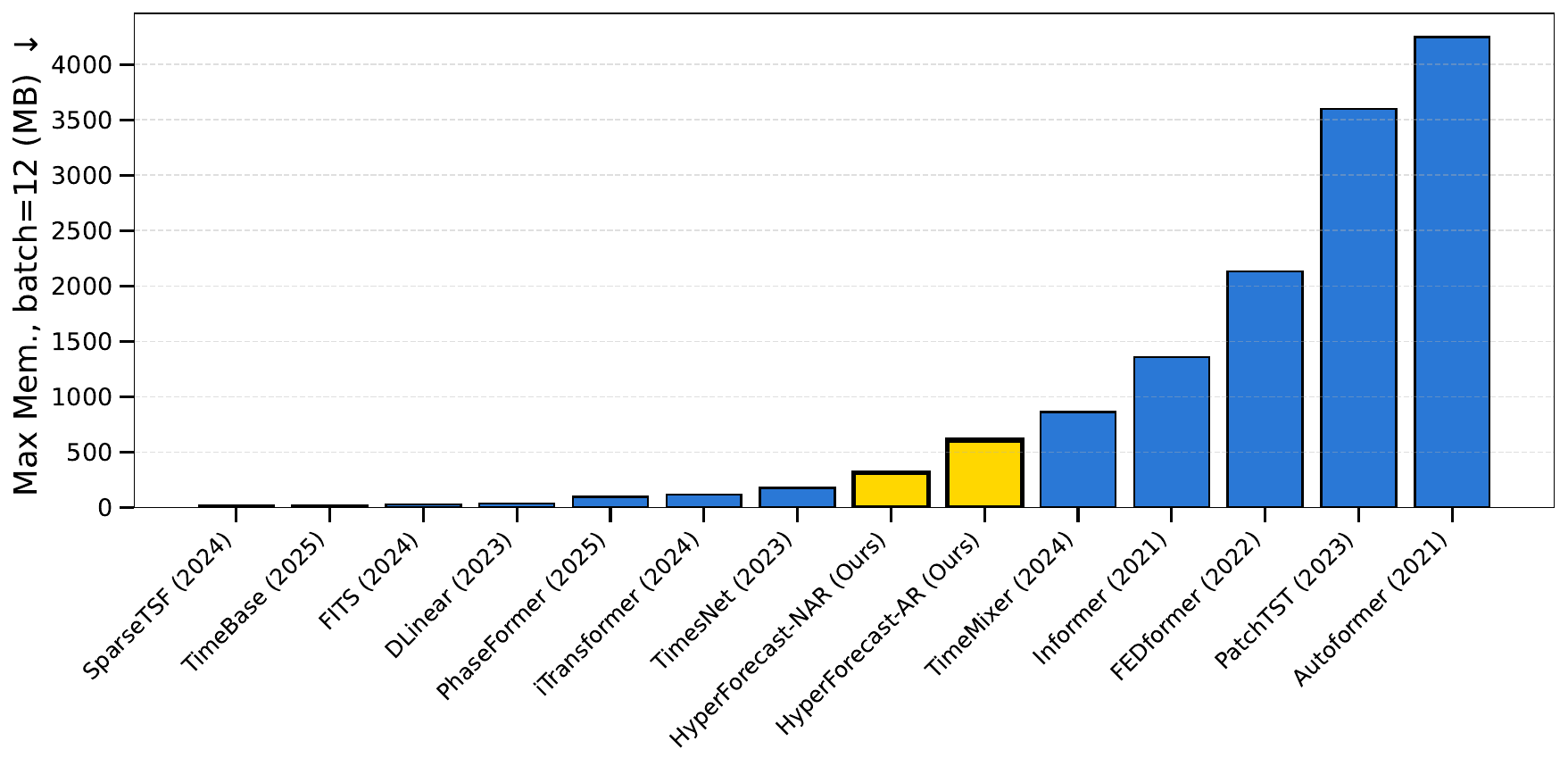}
        \caption{Peak memory at batch size 12, sorted ascending.}
        \label{fig:maxmem}
    \end{subfigure}
        \caption{Efficiency comparison across all fourteen models on Weather ($\text{seq\_len}=\text{pred\_len}=720$). \textbf{(a)} Parameter count vs.\ MACs for a single sample. \textbf{(b)} Peak memory at batch size 12, sorted ascending.}
    \label{fig:efficiency_combined}
\end{figure}

As shown in Figure~\ref{fig:efficiency}, HyperForecast occupies a distinct efficiency regime among the compared methods. Against the Transformer-based baselines (Informer, Autoformer, FEDformer, PatchTST), HyperForecast uses $264\times$ to $470\times$ fewer MACs (26.69M versus 7.05G to 12.62G) and up to $197\times$ fewer parameters (57.69K versus 11.38M for Informer), while remaining only slightly above DLinear (21.79M MACs) among the ultra-lightweight linear baselines. HyperForecast is therefore substantially lighter than any competitive Transformer-based method, at a small and bounded MAC cost relative to the linear family in exchange for its hierarchical hyperbolic representation. As Figure~\ref{fig:maxmem} shows, this advantage does not extend uniformly to measured peak memory. The segment-level hyperbolic operations used by HyperForecast, including logarithmic and exponential maps, parallel transport, and M\"obius arithmetic, retain more intermediate activation state per sample than a single linear layer, placing HyperForecast in the middle of the memory ranking rather than at the low end.

\begin{figure}[H]
    \centering
    \includegraphics[width=\linewidth]{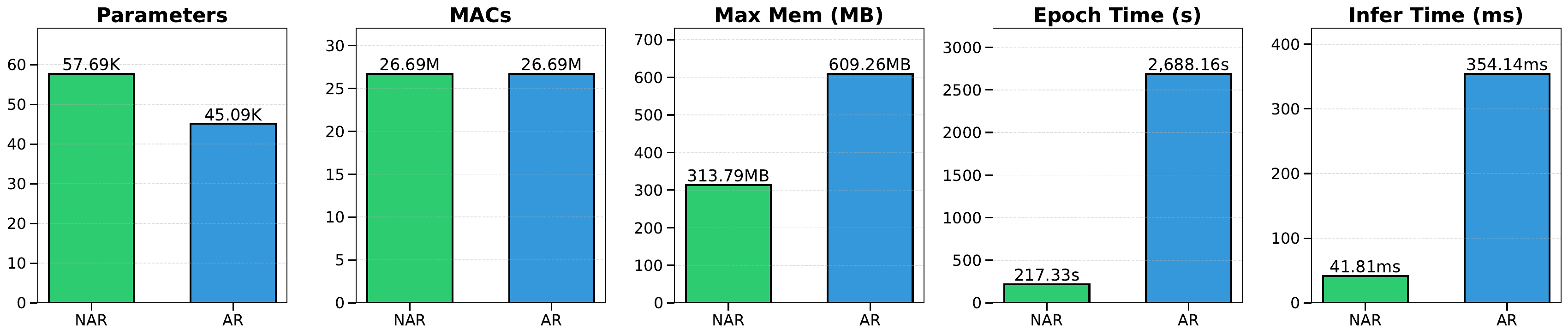}
        \caption{Efficiency of the two HyperForecast variants on Weather, seq\_len/pred\_len $=720/720$. NAR is shown in green and AR in blue.}
    \label{fig:nar_ar_tradeoff}
\end{figure}

In Figure~\ref{fig:nar_ar_tradeoff}, we isolate the cost of autoregressive decoding. As described in Sections~\ref{sec:ar} and~\ref{sec:nar}, the two variants share the same per-step machinery, namely geodesic velocity estimation through parallel-transported logarithmic maps (Eq.~\ref{eq:ar-vtilde}) followed by a Poincar\'e residual update. They differ in how this velocity is used and how often the machinery is invoked. NAR aggregates the velocity once over the full lookback and produces every future segment with a single batched call to $\exp_{\mathbf{z}_N}$ (Eq.~\ref{eq:nar-evolve}). AR instead re-estimates the velocity and re-applies the Poincar\'e residual update (Eq.~\ref{eq:ar-update}) once per predicted segment, and each step's moving window depends on the previous step's own output. This dependency makes the $H_{\text{pred}}$ steps inherently sequential and prevents them from being batched across time, the same structural bottleneck that makes autoregressive decoding slower than non-autoregressive decoding in sequence generation more broadly, compounded here by each recurrent step performing a nontrivial manifold operation rather than a single matrix multiply. This yields a $12.4\times$ epoch-time gap (2688.16s versus 217.33s) and an $8.5\times$ CPU inference-time gap (354.14ms versus 41.81ms) between AR and NAR, along with a smaller $1.9\times$ increase in peak memory (609.26MB versus 313.79MB) from retaining per-step intermediate state across the rollout. The two variants report identical MACs and FLOPs (26.69M and 53.38M) because our profiler traces a single forward call through the shared step function without multiplying by the number of sequential invocations. The MACs and FLOPs columns should therefore be read as per-step compute cost, while epoch time and inference time correctly reflect the $H_{\text{pred}}$-fold repetition that AR incurs at runtime. We report both variants for completeness and recommend NAR as the efficiency-preferred configuration, whereas AR remains useful where the recurrent error-correction rollout discussed in Section~\ref{sec:nar} is preferred despite its cost.

\section{Additional Spectral-Centroid Visualizations}
\label{app:centroid-figures}

As established in Figure~\ref{fig:centroid-solar} in Sec.~\ref{sec:radial_hierarchy}, the raw decomposed signal exhibits the target frequency ordering $T\!<\!S_c\!<\!S_f\!<\!R$ on Solar at $H=96$. Here we extend this evidence to four further representative benchmarks that include Weather, AQWan, METR-LA, and PM2.5, spanning channel counts from 11 to 207 and four distinct sensing domains (weather, air quality, traffic speed, and particulate matter).

\begin{figure*}[h]
\centering
\includegraphics[width=0.85\textwidth]{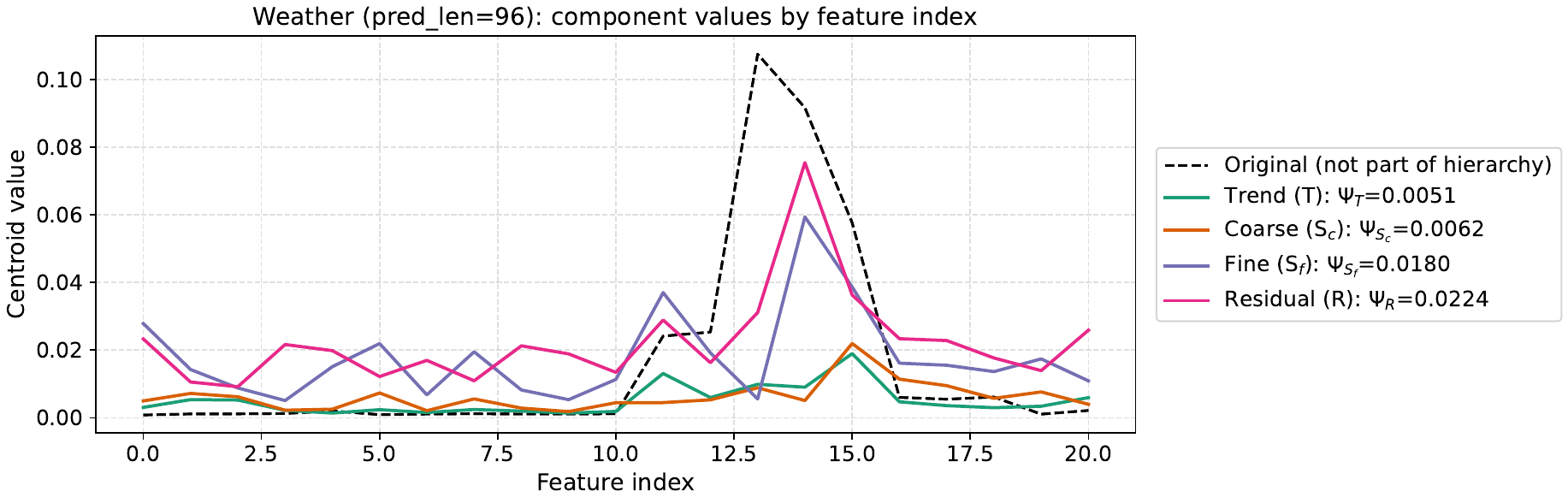}
\caption{Weather (21 channels, $H=96$). The full ordering holds on average centroids: $\Psi_T\!=\!0.0051 < \Psi_{S_c}\!=\!0.0062 < \Psi_{S_f}\!=\!0.0180 < \Psi_R\!=\!0.0224$.}
\label{fig:centroid-weather}
\end{figure*}

\begin{figure*}[h]
\centering
\includegraphics[width=0.85\textwidth]{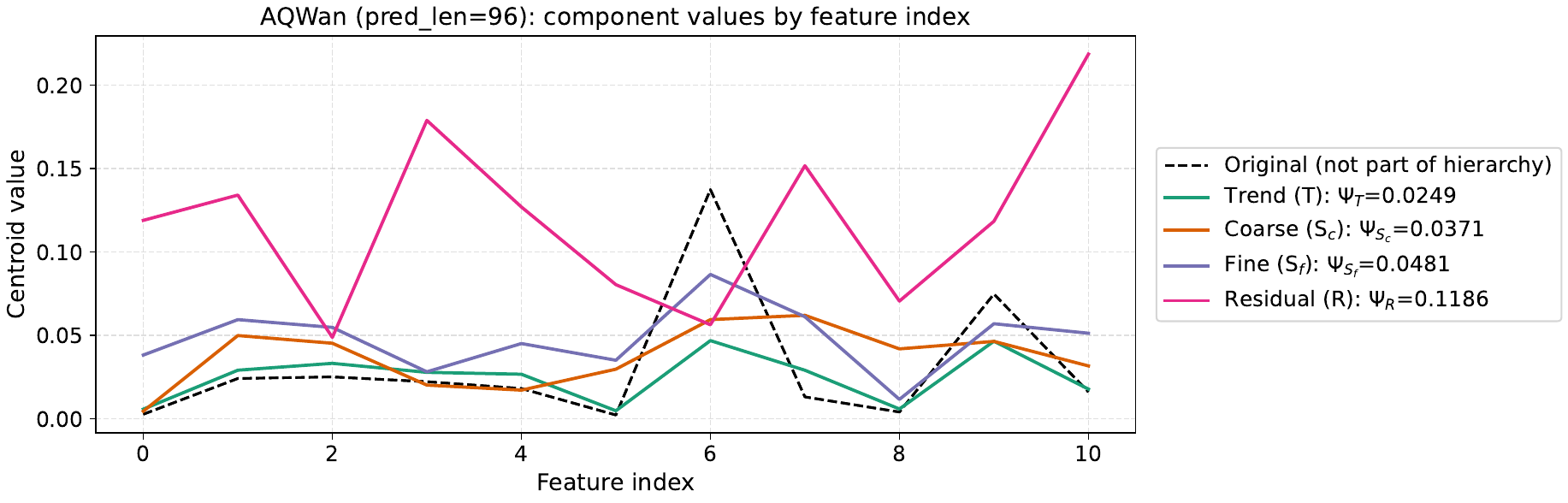}
\caption{AQWan (11 channels, $H=96$). The full ordering holds on average centroids: $\Psi_T\!=\!0.0249 < \Psi_{S_c}\!=\!0.0371 < \Psi_{S_f}\!=\!0.0481 < \Psi_R\!=\!0.1186$.}
\label{fig:centroid-aqwan}
\end{figure*}

\begin{figure*}[h]
\centering
\includegraphics[width=0.85\textwidth]{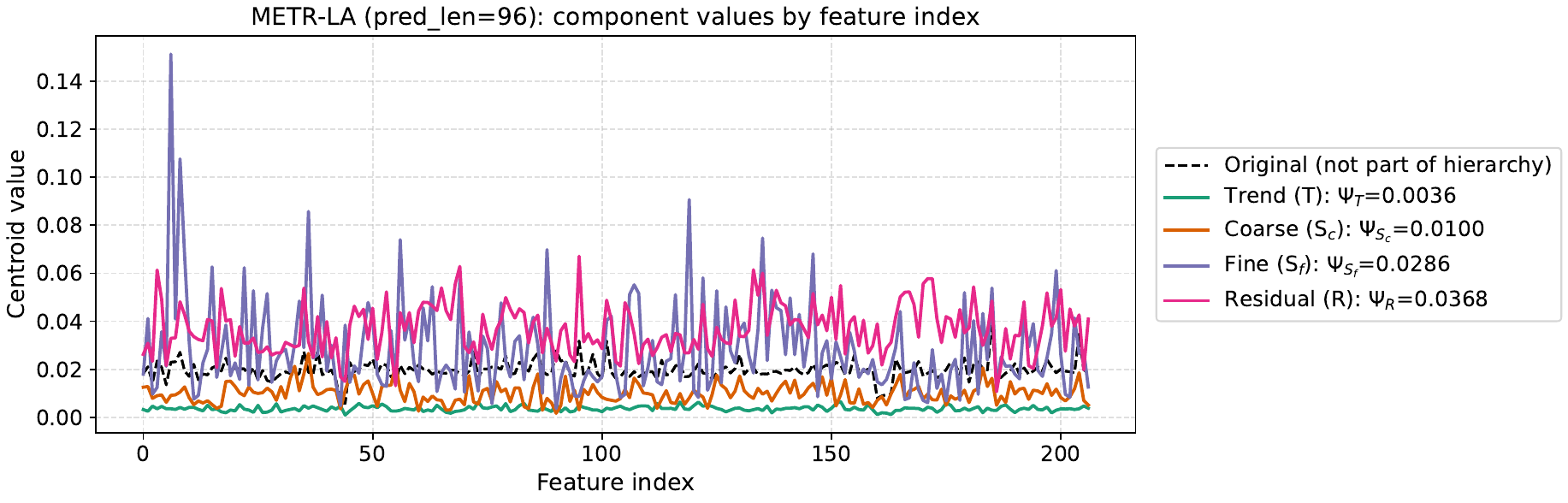}
\caption{METR-LA (207 channels, $H=96$). The full ordering holds on average centroids: $\Psi_T\!=\!0.0036 < \Psi_{S_c}\!=\!0.0100 < \Psi_{S_f}\!=\!0.0286 < \Psi_R\!=\!0.0368$.}
\label{fig:centroid-metrla}
\end{figure*}

\begin{figure*}[h]
\centering
\includegraphics[width=0.85\textwidth]{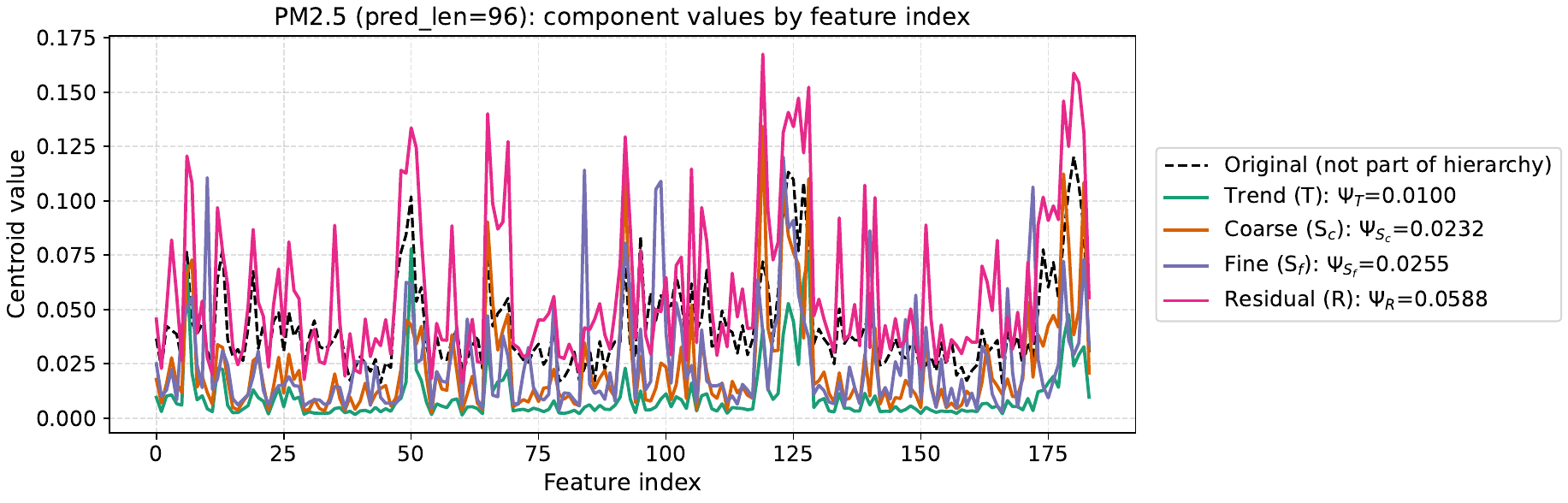}
\caption{PM2.5 (184 channels, $H=96$). The full ordering holds on average centroids: $\Psi_T\!=\!0.0100 < \Psi_{S_c}\!=\!0.0232 < \Psi_{S_f}\!=\!0.0255 < \Psi_R\!=\!0.0588$.}
\label{fig:centroid-pm25}
\end{figure*}

Across these benchmarks, spanning different channel range and domains, the cascaded decomposition separates raw-signal frequency content in the same direction, $T\!<\!S_c\!<\!S_f\!<\!R$. This consistency, obtained without any hyperbolic embedding or hierarchy loss in the loop, supports treating the frequency hierarchy as a property of the decomposition itself rather than an artifact of training on any one domain. Moreover, Appendix~\ref{app:hierloss-ablation} provides complementary evidence that the corresponding radial ordering in the trained embedding persists even when the hierarchy loss is removed.

\section{Ablation Studies}
\label{sec:ablations}
\subsection{Impact of Learnable Decomposition}

Our learnable decomposition module employs cascaded depthwise convolutions with component-specific kernel sizes to extract trend, coarse seasonal, fine seasonal, and residual components from raw input sequences. To isolate the contribution of Learnable Decomposition, we compare our approach against a variant of our model that does not implement Learnable Decomposition; it merely processes raw time series directly through segment encoding and hyperbolic dynamics. 

\begin{table}[h]
\centering
\scriptsize
\caption{Impact of learnable decomposition on forecasting accuracy, NAR and AR. Models with decomposition (trend, coarse/fine seasonal, residual) vs. without decomposition across prediction horizons $H \in \{96, 192, 336, 720\}$.}
\label{tab:ablation_decomposition}
\resizebox{0.85\columnwidth}{!}{
\begin{tabular}{llcccccccc}
\toprule
Dataset & Variant
& \multicolumn{2}{c}{96} & \multicolumn{2}{c}{192} & \multicolumn{2}{c}{336} & \multicolumn{2}{c}{720} \\
\cmidrule(lr){3-4} \cmidrule(lr){5-6} \cmidrule(lr){7-8} \cmidrule(lr){9-10}
& & MSE & MAE & MSE & MAE & MSE & MAE & MSE & MAE \\
\midrule
\multirow{4}{*}{AQShunyi}
& NAR w/o Decomp. & 0.638 & 0.490 & 0.679 & 0.507 & 0.713 & 0.527 & 0.776 & 0.554 \\
& NAR w/ Decomp. & \textbf{0.607} & \textbf{0.475} & \textbf{0.640} & \textbf{0.498} & \textbf{0.670} & \textbf{0.510} & \textbf{0.712} & \textbf{0.528} \\
\cmidrule(lr){2-10}
& AR w/o Decomp.  & 0.631 & 0.483 & 0.690 & 0.512 & 0.713 & 0.524 & 0.773 & 0.551 \\
& AR w/ Decomp.   & \textbf{0.606} & \textbf{0.473} & \textbf{0.642} & \textbf{0.493} & \textbf{0.669} & \textbf{0.508} & \textbf{0.742} & \textbf{0.542} \\
\midrule
\multirow{4}{*}{AQWan}
& NAR w/o Decomp. & 0.717 & 0.471 & 0.779 & 0.508 & 0.802 & 0.514 & 0.876 & 0.547 \\
& NAR w/ Decomp. & \textbf{0.692} & \textbf{0.458} & \textbf{0.733} & \textbf{0.479} & \textbf{0.760} & \textbf{0.492} & \textbf{0.819} & \textbf{0.514} \\
\cmidrule(lr){2-10}
& AR w/o Decomp.  & 0.713 & 0.470 & 0.776 & 0.501 & 0.805 & 0.514 & 0.875 & 0.541 \\
& AR w/ Decomp.   & \textbf{0.696} & \textbf{0.464} & \textbf{0.742} & \textbf{0.486} & \textbf{0.766} & \textbf{0.499} & \textbf{0.846} & \textbf{0.532} \\
\midrule
\multirow{4}{*}{Weather}
& NAR w/o Decomp. & 0.188 & 0.243 & 0.218 & 0.271 & 0.268 & 0.309 & 0.333 & 0.355 \\
& NAR w/ Decomp. & \textbf{0.145} & \textbf{0.206} & \textbf{0.188} & \textbf{0.252} & \textbf{0.237} & \textbf{0.292} & \textbf{0.303} & \textbf{0.344} \\
\cmidrule(lr){2-10}
& AR w/o Decomp.  & 0.167 & 0.220 & 0.213 & 0.263 & 0.274 & 0.318 & 0.339 & 0.361 \\
& AR w/ Decomp.   & \textbf{0.143} & \textbf{0.202} & \textbf{0.190} & \textbf{0.248} & \textbf{0.238} & \textbf{0.289} & \textbf{0.315} & \textbf{0.341} \\
\bottomrule
\end{tabular}
}
\end{table}

Results in Table~\ref{tab:ablation_decomposition} demonstrate that decomposition effectiveness is dataset- and horizon-dependent, and the pattern holds for both forecasting variants. On Weather, with its strong short-range multi-frequency structure (daily/weekly cycles across 21 channels), the NAR gain is largest at short horizons (22.9\% MSE reduction at $H=96$) and tapers to 9.0\% at $H=720$. AR shows the same shape (14.4\% at $H=96$ tapering to 7.1\% at $H=720$). Air quality datasets (AQShunyi, AQWan) show steady 3.5--4.9\% NAR gains at $H=96$, growing to 6.5--8.3\% at $H=720$; AR gains are similar in magnitude (2.4--4.0\% at $H=96$, 3.3--4.0\% at $H=720$), suggesting decomposition's value there comes more from stabilizing long-horizon rollout than from separating short-term seasonal content, for both variants.

\subsection{Channel-Independent vs.\ Channel-Dependent Processing}
\label{app:ci-cd}
Our original methodology processes each channel independently, folding the
feature dimension into the batch axis so that a shared encoder sees $C$
univariate sequences. We compare against a channel-dependent variant that
instead keeps all $C$ channels in a single joint tensor through the encoder,
on Solar, AQShunyi, and AQWan (Poincar\'e), for both the NAR and AR variants.

\begin{table}[h]
\centering
\scriptsize
\caption{Channel-independent vs.\ channel-dependent processing (Poincar\'e, NAR; MSE).}
\label{tab:ci-cd}
\begin{tabular}{llcccc}
\toprule
Dataset & Variant & 96 & 192 & 336 & 720 \\
\midrule
\multirow{2}{*}{Solar}    & Channel-Dependent   & 0.207 & 0.234 & 0.239 & 0.223 \\
                          & Channel-Independent & \textbf{0.177} & \textbf{0.204} & \textbf{0.221} & \textbf{0.217} \\
\midrule
\multirow{2}{*}{AQShunyi} & Channel-Dependent   & 0.667 & 0.678 & 0.693 & \textbf{0.689} \\
                          & Channel-Independent & \textbf{0.607} & \textbf{0.646} & \textbf{0.678} & 0.717 \\
\midrule
\multirow{2}{*}{AQWan}    & Channel-Dependent   & 0.774 & 0.754 & \textbf{0.770} & 0.829 \\
                          & Channel-Independent & \textbf{0.727} & \textbf{0.741} & 0.773 & \textbf{0.822} \\
\bottomrule
\end{tabular}
\end{table}

Channel-independent processing wins on 10 of 12 dataset/horizon cells, and wins all four horizons on Solar. The two exceptions, AQShunyi at $H=720$ and AQWan at $H=336$, differ in margin (4.1\% and 0.4\% relative, respectively) at different horizons in each dataset, so we don't read a systematic horizon-dependent crossover into them. Channel-independent processing is the consistently better choice across the three datasets tested. We use it as the default given its advantage at the horizons where HyperForecast shows its largest gains (Table~\ref{tab:main_results_mean}).

\begin{table}[h]
\centering
\scriptsize
\caption{Channel-independent vs.\ channel-dependent processing (Poincar\'e, AR; MSE).}
\label{tab:ci-cd-ar}
\begin{tabular}{llcccc}
\toprule
Dataset & Variant & 96 & 192 & 336 & 720 \\
\midrule
\multirow{2}{*}{Solar}    & Channel-Dependent   & 0.206 & 0.738 & 0.754 & 0.757 \\
                          & Channel-Independent & \textbf{0.170} & \textbf{0.206} & \textbf{0.227} & \textbf{0.243} \\
\midrule
\multirow{2}{*}{AQShunyi} & Channel-Dependent   & 0.658 & 0.684 & 0.692 & \textbf{0.713} \\
                          & Channel-Independent & \textbf{0.615} & \textbf{0.658} & \textbf{0.690} & 0.756 \\
\midrule
\multirow{2}{*}{AQWan}    & Channel-Dependent   & 0.757 & 0.765 & 0.787 & \textbf{0.839} \\
                          & Channel-Independent & \textbf{0.704} & \textbf{0.754} & \textbf{0.781} & 0.867 \\
\bottomrule
\end{tabular}
\end{table}

The AR variant shows the same pattern as NAR, channel-independent processing wins 10 of 12 dataset/horizon cells, with both exceptions at $H=720$ (AQShunyi and AQWan), where channel-dependent's larger joint parameterization appears to help stabilize the longest, error-accumulating AR rollout. The channel-dependent AR variant is also markedly less stable on Solar (137 channels) beyond $H=96$, where its MSE roughly triples rather than growing gracefully with horizon, consistent with a joint 137-channel tensor being harder for autoregressive rollout to fit than the channel-independent formulation. Taken together, channel-independent processing is the more robust default across both forecasting paradigms.

\subsection{AR Window Size Sensitivity}
\label{app:window-size}

The AR branch aggregates a moving window of $K$ historical segments into each rollout-step velocity estimate (Eq.~\ref{eq:ar-vtilde}). We sweep $K\in\{2,4,5,6\}$ on AQShunyi, Temp, and ZafNoo.
 
\begin{table}[h]
\centering
\caption{AR window size $K$ ablation (MSE). Best per dataset/horizon in \textbf{bold}.}
\label{tab:window-size}
\scriptsize
\begin{tabular}{llcccc}
\toprule
Dataset & $K$ & 96 & 192 & 336 & 720 \\
\midrule
\multirow{4}{*}{AQShunyi} & 2 & 0.615 & \textbf{0.658} & 0.690 & 0.756 \\
                          & 4 & 0.617 & 0.662 & \textbf{0.689} & 0.758 \\
                          & 5 & \textbf{0.613} & 0.661 & 0.691 & 0.758 \\
                          & 6 & 0.616 & 0.662 & \textbf{0.689} & \textbf{0.737} \\
\midrule
\multirow{4}{*}{Temp}     & 2 & 0.126 & 0.125 & 0.136 & 0.637 \\
                          & 4 & 0.124 & 0.126 & 0.138 & 0.293 \\
                          & 5 & 0.124 & \textbf{0.122} & 0.150 & 0.407 \\
                          & 6 & \textbf{0.122} & 0.166 & \textbf{0.130} & \textbf{0.203} \\
\midrule
\multirow{4}{*}{ZafNoo}   & 2 & 0.422 & 0.481 & 0.526 & 0.579 \\
                          & 4 & \textbf{0.419} & 0.479 & 0.514 & 0.580 \\
                          & 5 & \textbf{0.419} & 0.472 & 0.519 & \textbf{0.566} \\
                          & 6 & 0.420 & \textbf{0.472} & \textbf{0.508} & 0.571 \\
\bottomrule
\end{tabular}
\end{table}
 
At $H\leq336$, sensitivity to $K$ is modest on AQShunyi and ZafNoo (under 3.54\% MSE spread across $K\in\{2,4,5,6\}$), but Temp already shows substantial sensitivity at these horizons (up to 36\% spread at $H=192$). At $H=720$, $K$ interacts strongly with dataset-specific dynamics: on Temp, $K=2$ is markedly unstable (MSE 0.637, over $3\times$ worse than $K=6$'s 0.203), while AQShunyi and ZafNoo show only mild sensitivity at the same horizon (under 2.85\% spread). This indicates Temp's velocity field benefits from a wider historical window specifically at long rollout horizons, where a narrow 2-segment window is more exposed to compounding local noise across many rollout steps. $K\in\{4,6\}$ is a robust default, and $K=2$ should be avoided for long-horizon AR forecasting on datasets with Temp-like dynamics.
 
\subsection{Impact of Auxiliary Losses}
\label{app:loss-ablation}

To isolate the necessity of the two auxiliary losses (as opposed to their weight sensitivity, below), we retrain HyperForecast NAR on CzeLan, AQWan, and Exchange with the temporal-consistency loss removed, the hierarchy loss removed, and both removed, holding all other hyperparameters fixed.
 
\begin{table}[h]
\centering
\scriptsize
\caption{Effect of removing the temporal-consistency loss $\mathcal{L}_{\text{tc}}$ and hierarchy loss $\mathcal{L}_{\text{hier}}$ on HyperForecast (NAR), per horizon. Best per dataset/horizon in \textbf{bold}.}
\label{tab:loss-ablation}
\begin{tabular}{llcccc}
\toprule
Dataset & Variant & 96 & 192 & 336 & 720 \\
\midrule
\multirow{4}{*}{CzeLan}   & Main       & \textbf{0.171} & 0.205          & \textbf{0.237} & \textbf{0.270} \\
                          & w/o TC     & 0.183          & \textbf{0.203} & 0.239          & 0.331 \\
                          & w/o Hier   & 0.190          & 0.223          & 0.240          & 0.279 \\
                          & w/o both   & 0.200          & 0.211          & 0.238          & 0.285 \\
\midrule
\multirow{4}{*}{AQWan}    & Main       & \textbf{0.692} & \textbf{0.733} & \textbf{0.760} & 0.819 \\
                          & w/o TC     & 0.699          & 0.735          & 0.764          & \textbf{0.815} \\
                          & w/o Hier   & 0.701          & 0.746          & 0.773          & 0.825 \\
                          & w/o both   & 0.703          & 0.739          & 0.774          & 0.823 \\
\midrule
\multirow{4}{*}{Exchange} & Main       & \textbf{0.079} & 0.165          & \textbf{0.305} & \textbf{0.607} \\
                          & w/o TC     & 0.080          & \textbf{0.163} & 0.745          & 0.752 \\
                          & w/o Hier   & 0.120          & 0.210          & 0.558          & 1.910 \\
                          & w/o both   & 0.124          & 0.209          & 0.528          & 1.767 \\
\bottomrule
\end{tabular}
\end{table}
 
Removing either loss degrades performance on all three datasets, and the effect is far larger on Exchange (main 0.289 $\to$ 0.700 MSE without $\mathcal{L}_{\text{hier}}$, +142\%) than on CzeLan or AQWan (both under 6\% degradation). Exchange is also the shortest and lowest-channel-count series of the three (Table~\ref{tab:datasets}, 8 channels, 7{,}588 timesteps versus CzeLan's 11 and 19{,}934 and AQWan's 11 and 35{,}064), giving the architecture less raw signal to passively exploit for frequency separation, so its embedding geometry likely depends more heavily on the explicit hierarchy constraint, consistent with the finding in Appendix~\ref{app:hierloss-ablation} that the hierarchy loss sharpens, rather than creates, the radial ordering.
 
\subsection{Temporal Consistency and Hierarchy Losses}
\label{app:loss-weight-sensitivity}
HyperForecast employs two auxiliary losses: temporal consistency loss $\mathcal{L}_{\text{tc}}$ penalizes abrupt hyperbolic velocity changes between prediction steps, while hierarchy loss $\mathcal{L}_{\text{hier}}$ enforces radial ordering (trend near origin, residual near boundary). We perform a grid search over both loss weights on ZafNoo, Temp, and AQShunyi to identify optimal regularization strengths (all other hyperparameters fixed, $H=96$).

\begin{table}[h]
\centering
\scriptsize
\caption{Hierarchy-loss weight sensitivity at $H=96$ (NAR), with $\lambda_{\text{tc}}=0.1$ fixed. MSE reported; best per dataset in \textbf{bold}.}
\label{tab:loss-weight-sensitivity-hier}
\begin{tabular}{lccccccc}
\toprule
& \multicolumn{7}{c}{$\lambda_{\text{hier}}$} \\
\cmidrule(lr){2-8}
Dataset & 0 & 0.0001 & 0.001 & 0.005 & 0.01 & 0.05 & 0.1 \\
\midrule
ZafNoo   & 0.439 & 0.429 & 0.429 & 0.425 & 0.423 & \textbf{0.419} & 0.422 \\
Temp     & 0.128 & 0.128 & 0.135 & \textbf{0.124} & 0.124 & 0.128 & 0.138 \\
AQShunyi & 0.610 & \textbf{0.605} & 0.617 & 0.619 & 0.619 & 0.615 & 0.609 \\
\bottomrule
\end{tabular}
\end{table}
 
\begin{table}[ht]
\centering
\scriptsize
\caption{Temporal-consistency loss weight sensitivity at $H=96$ (NAR), with $\lambda_{\text{hier}}=0.01$ fixed. MSE reported; best per dataset in \textbf{bold}.}
\label{tab:loss-weight-sensitivity-tc}
\begin{tabular}{lcccc}
\toprule
& \multicolumn{4}{c}{$\lambda_{\text{tc}}$} \\
\cmidrule(lr){2-5}
Dataset & 0 & 0.05 & 0.1 & 0.2 \\
\midrule
ZafNoo   & 0.425 & 0.425 & \textbf{0.423} & 0.430 \\
Temp     & 0.125 & \textbf{0.124} & 0.124 & 0.124 \\
AQShunyi & 0.619 & \textbf{0.618} & 0.619 & 0.618 \\
\bottomrule
\end{tabular}
\end{table}
 
The optimal hierarchy weight is dataset-dependent: ZafNoo prefers the strongest setting tested ($\lambda_{\text{hier}}=0.05$, 4.6\% MSE reduction over $\lambda_{\text{hier}}=0$), Temp prefers a moderate weight ($0.005$, 3.5\% reduction), and AQShunyi prefers the weakest nonzero weight ($0.0001$, 0.9\% reduction). Removing the hierarchy loss entirely ($\lambda_{\text{hier}}=0$) is worst or near-worst on all three, most clearly on ZafNoo, consistent with the necessity result above, where removing $\mathcal{L}_{\text{hier}}$ also hurt every dataset tested. Sensitivity to the temporal-consistency weight is an order of magnitude smaller across all three datasets (differences under 1.7\% MSE between any two weights), indicating the model is comparatively robust to this hyperparameter and the hierarchy weight is the one worth tuning per dataset.
 
\section{Run-to-Run Stability}
\label{sec:stability}

To confirm the reported gains are not seed noise, we retrain HyperForecast NAR on seven representative datasets across 5 seeds. Standard deviations stay under 0.011 MSE on most datasets, with PM2.5 being a consistent exception, having MSE std between 0.010--0.027 across horizons (up to 6.5\% of its mean at $H=720$).
 
\begin{table*}[h]
\centering
\scriptsize
\caption{Mean $\pm$ std MSE/MAE over 5 seeds for HyperForecast NAR.}
\label{tab:stability}
\begin{tabular}{lcccc}
\toprule
Dataset & $H$ & MSE (mean $\pm$ std) & MAE (mean $\pm$ std) \\
\midrule
\multirow{4}{*}{Weather}  & 96  & 0.146 $\pm$ 0.002 & 0.210 $\pm$ 0.003 \\
                          & 192 & 0.188 $\pm$ 0.001 & 0.253 $\pm$ 0.002 \\
                          & 336 & 0.245 $\pm$ 0.010 & 0.301 $\pm$ 0.012 \\
                          & 720 & 0.309 $\pm$ 0.005 & 0.349 $\pm$ 0.006 \\
\midrule
\multirow{4}{*}{AQShunyi} & 96  & 0.609 $\pm$ 0.003 & 0.474 $\pm$ 0.002 \\
                          & 192 & 0.646 $\pm$ 0.002 & 0.498 $\pm$ 0.005 \\
                          & 336 & 0.680 $\pm$ 0.008 & 0.518 $\pm$ 0.006 \\
                          & 720 & 0.718 $\pm$ 0.003 & 0.533 $\pm$ 0.004 \\
\midrule
\multirow{4}{*}{AQWan}    & 96  & 0.698 $\pm$ 0.001 & 0.465 $\pm$ 0.004 \\
                          & 192 & 0.738 $\pm$ 0.008 & 0.482 $\pm$ 0.003 \\
                          & 336 & 0.765 $\pm$ 0.003 & 0.493 $\pm$ 0.003 \\
                          & 720 & 0.821 $\pm$ 0.008 & 0.520 $\pm$ 0.008 \\
\midrule
\multirow{4}{*}{ZafNoo}   & 96  & 0.420 $\pm$ 0.003 & 0.404 $\pm$ 0.004 \\
                          & 192 & 0.474 $\pm$ 0.006 & 0.441 $\pm$ 0.007 \\
                          & 336 & 0.517 $\pm$ 0.008 & 0.468 $\pm$ 0.006 \\
                          & 720 & 0.554 $\pm$ 0.011 & 0.481 $\pm$ 0.007 \\
\midrule
\multirow{4}{*}{PM2.5}    & 96  & 0.428 $\pm$ 0.011 & 0.452 $\pm$ 0.014 \\
                          & 192 & 0.425 $\pm$ 0.010 & 0.453 $\pm$ 0.015 \\
                          & 336 & 0.428 $\pm$ 0.018 & 0.478 $\pm$ 0.021 \\
                          & 720 & 0.414 $\pm$ 0.027 & 0.487 $\pm$ 0.023 \\
\midrule
\multirow{4}{*}{Temp}     & 96  & 0.126 $\pm$ 0.002 & 0.273 $\pm$ 0.003 \\
                          & 192 & 0.123 $\pm$ 0.001 & 0.272 $\pm$ 0.001 \\
                          & 336 & 0.118 $\pm$ 0.001 & 0.266 $\pm$ 0.001 \\
                          & 720 & 0.119 $\pm$ 0.006 & 0.267 $\pm$ 0.007 \\
\midrule
\multirow{4}{*}{Wind}     & 96  & 0.774 $\pm$ 0.005 & 0.621 $\pm$ 0.004 \\
                          & 192 & 0.905 $\pm$ 0.009 & 0.689 $\pm$ 0.003 \\
                          & 336 & 1.007 $\pm$ 0.006 & 0.742 $\pm$ 0.001 \\
                          & 720 & 1.051 $\pm$ 0.006 & 0.773 $\pm$ 0.003 \\
\bottomrule
\end{tabular}
\end{table*}
 
\section{Additional Hyperbolic Embedding Visualizations}
\label{app:additional_embeddings}
\begin{figure*}[h]
\centering
\includegraphics[width=\textwidth]{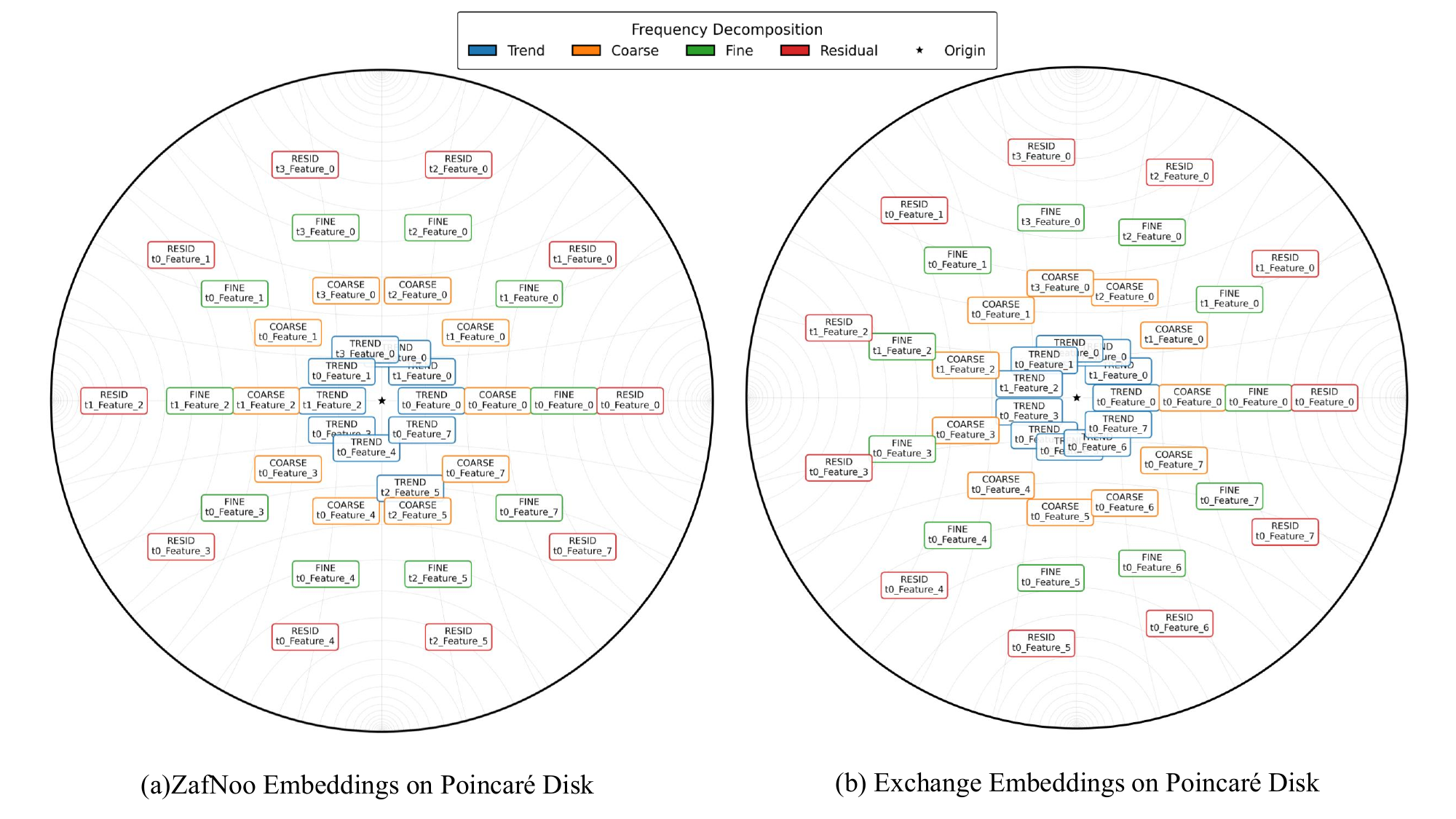}
\caption{Each point represents a (Scale\_TimeSegment\_Feature) tuple at a specific frequency scale. \textcolor{blue}{Trend} (innermost) captures low-frequency patterns closest to the origin, while \textcolor{red}{Residual} (outermost) encodes high-frequency noise farther from the origin. The \textbf{radial hierarchy} demonstrates that our model successfully learns frequency-based decomposition (Trend $<$ Coarse $<$ Fine $<$ Residual). Node labels follow the format \texttt{t\{segment\}\_\{feature\}} (e.g., t0\_Feature\_1 = time segment 0, feature 1).}
\label{fig:plot3}
\end{figure*}

To further validate the consistency of learned frequency-based hierarchies across diverse datasets, we provide Poincar\'e disk visualizations for two additional benchmarks with markedly different temporal characteristics, ZafNoo (an evapotranspiration dataset with strong multi-scale seasonality) and Exchange (a financial time series with weak periodic structure) (Figure~\ref{fig:plot3}). The autoregressive (AR) forecasting and non-autoregressive (NAR) variant produce embeddings with similar hierarchical structure on both datasets.

The fact that both datasets, despite very different periodicity profiles, produce the same hierarchical structure validates that hyperbolic geometry adapts to varying degrees of temporal complexity and that the decomposition reliably induces frequency-based radial separation across diverse domains.

\section{Qualitative Geometry Comparison: Poincar\'e vs.\ Euclidean Embeddings}
\label{app:geometry-vs-euclidean}

\begin{figure}[h]
    \centering
    \includegraphics[width=\linewidth]{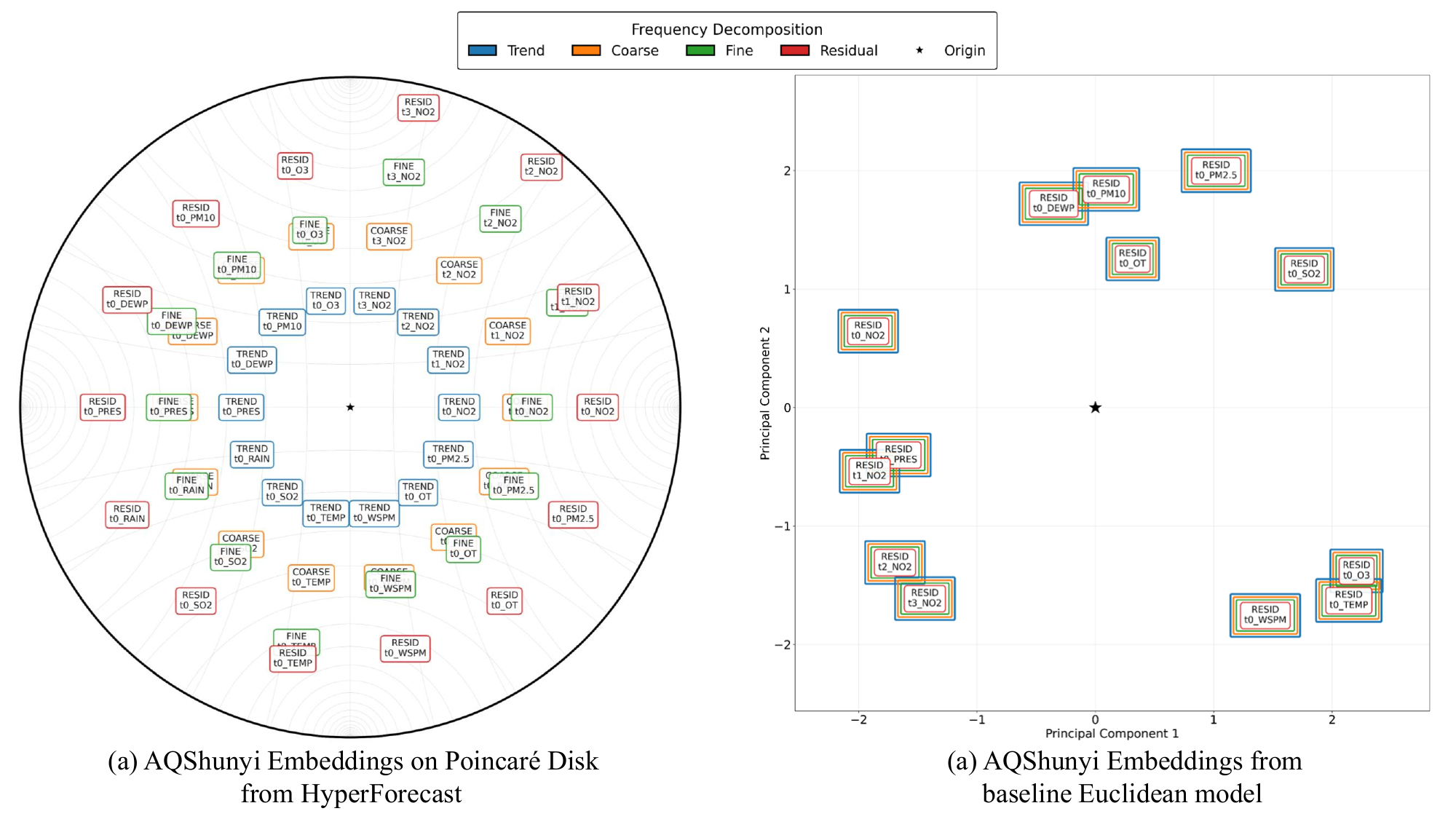}
    \caption{Segment embeddings learned on AQShunyi.
    \textbf{Left:} Poincar\'e disk from HyperForecast.
    Trend embeddings cluster near the origin, while coarse, fine, and residual components sit at increasing radii.
    \textbf{Right:} Euclidean PCA from the same architecture. All four frequency components collapse to coincident 2D positions, eliminating the radial hierarchy.}
    \label{fig:emb-aqshunyi}
\end{figure}

\begin{figure}[H]
    \centering
    \includegraphics[width=\linewidth]{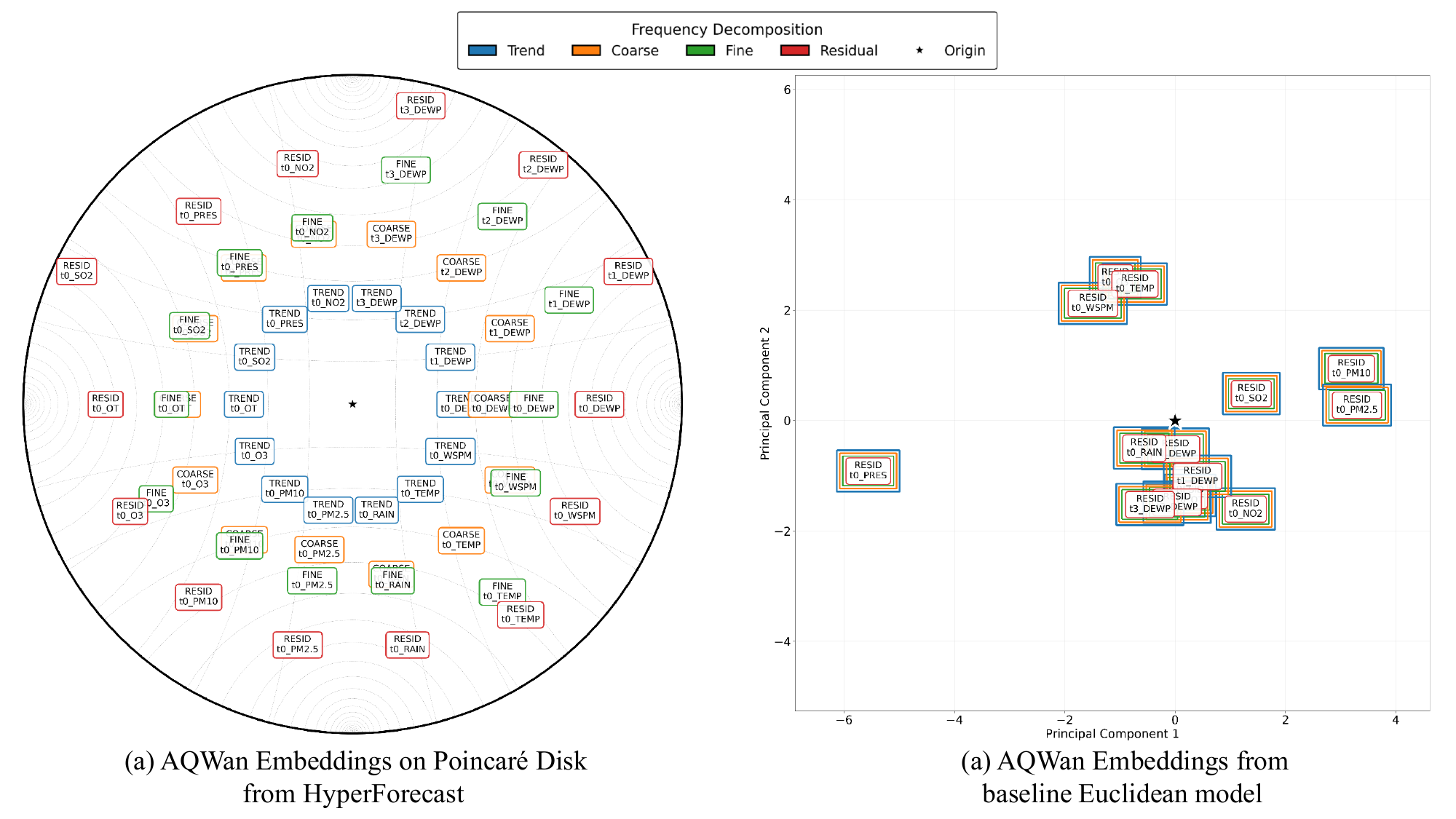}
    \caption{Segment embeddings learned on AQWan; conventions follow
    Figure~\ref{fig:emb-aqshunyi}. Similar radial ordering emerges in the Poincar\'e panel and is absent in the Euclidean PCA, where the four components stack at identical coordinates.}
    \label{fig:emb-aqwan}
\end{figure}

To complement the quantitative geometry ablation in Sec.~\ref{sec:ablations}, we visualize the learned segment embeddings produced by HyperForecast (Poincar\'e) and by a controlled Euclidean variant of the same architecture on two representative air-quality benchmarks: AQShunyi (Figure~\ref{fig:emb-aqshunyi}) and AQWan (Figure~\ref{fig:emb-aqwan}). The Euclidean variant retains an identical learnable decomposition, segment encoder, RevIN normalization, and forecasting head, replacing only the hyperbolic dynamics with their flat counterparts. The Euclidean embeddings are projected to two dimensions via PCA; the Poincar\'e embeddings are visualized directly on the unit disk via UMAP.

The Poincar\'e panels (left in each figure) exhibit an unambiguous radial separation by frequency band. Trend embeddings cluster tightly near the origin, while coarse seasonal, fine seasonal, and residual components occupy progressively larger radii, realizing the ordering $r_{\mathbf{T}} < r_{\mathbf{S}_c} < r_{\mathbf{S}_f} < r_{\mathbf{R}}$ induced by the cascaded decomposition. In the Euclidean PCA panels (right), this structure collapses. For every $\langle$\textsc{timestep}, \textsc{feature}$\rangle$ pair, the four frequency bands project to nearly coincident 2D coordinates. The flat geometry, which allocates representational capacity uniformly across the Euclidean space, has no mechanism for encoding the frequency hierarchy, so decomposition is preserved as a \emph{label} but not as a \emph{geometric structure}. This visual collapse is consistent with the MSE degradation reported when the manifold is replaced (Sec.~\ref{sec:ablations}).

\section{Radial Hierarchy Emerges Without Hierarchical Constraint}
\label{app:hierloss-ablation}

A natural concern is whether the radial frequency hierarchy reported is a genuine property of the embedding geometry or merely an artefact of the hierarchical radial loss $\mathcal{L}_{\mathrm{hier}}$, which explicitly penalizes deviations from the target ordering. To rule out this confound, we retrain HyperForecast on AQWan with $\lambda_{\mathrm{hier}} = 0$ (no explicit hierarchy constraint) and compare the resulting Poincar\'e embeddings against the standard run with $\lambda_{\mathrm{hier}} = 1e-4$.

\begin{figure*}[h]
    \centering
    \includegraphics[width=\linewidth]{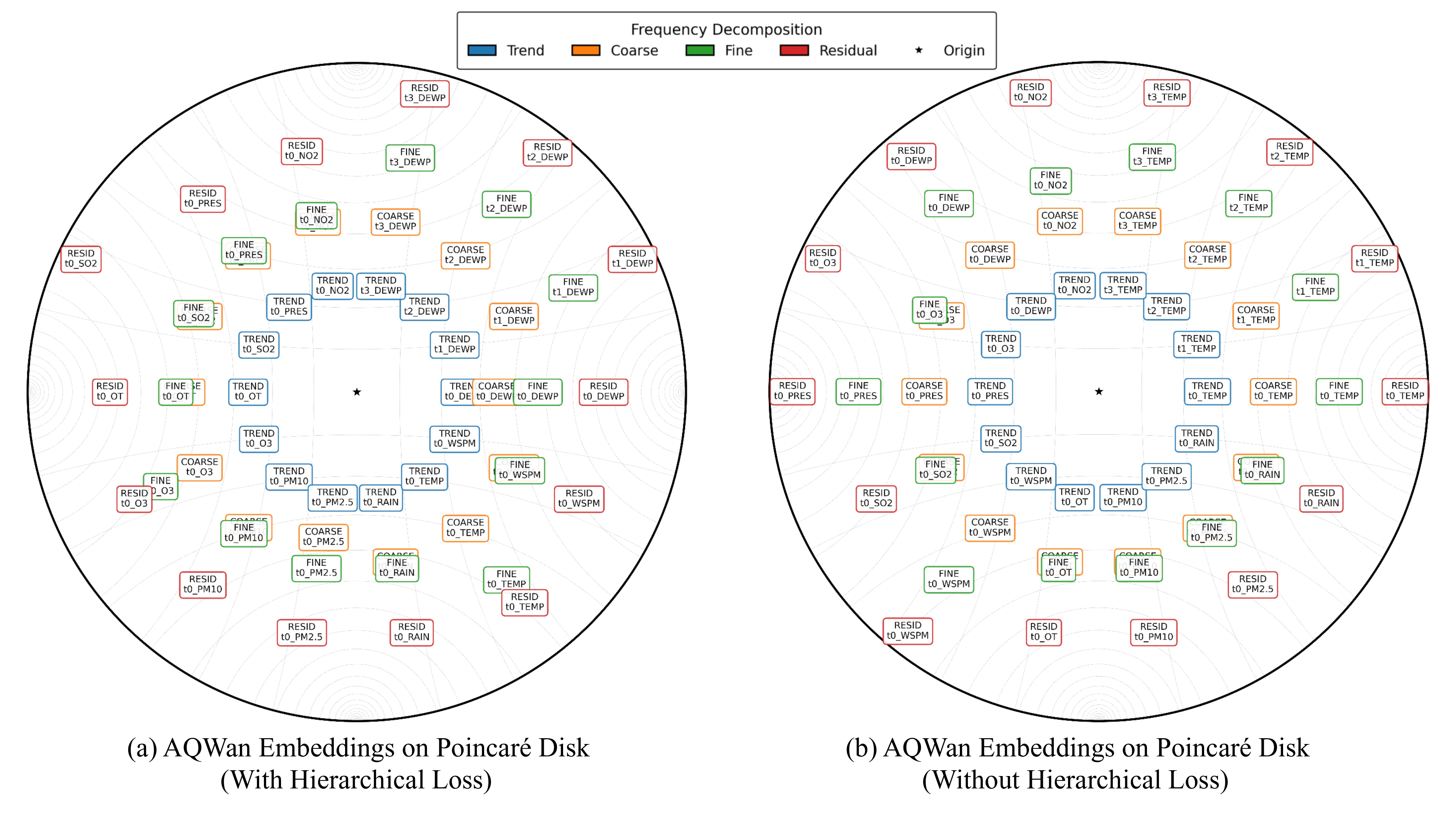}
    \caption{Poincar\'e-disk embeddings learned on AQWan with and without the hierarchical radial loss $\mathcal{L}_{\mathrm{hier}}$.}
    \label{fig:hierloss-ablation}
\end{figure*}

Both embeddings on the Poincar\'e disk are illustrated in Figure~\ref{fig:hierloss-ablation}. Both panels show the same monotonic ordering $r_{\mathbf{T}} < r_{\mathbf{S}_c} < r_{\mathbf{S}_f} < r_{\mathbf{R}}$. This is consistent with our design hypothesis, that the radial hierarchy is a geometric consequence of the interaction between the cascaded decomposition, which produces components of strictly decreasing temporal smoothness and strictly increasing variance, and the exponential volume growth of $\mathbb{P}^d_c$, which naturally allocates more capacity to higher-variance signals at larger radii. The hierarchical loss \emph{sharpens} an ordering that the architecture already prefers. We therefore report results with $\lambda_{\mathrm{hier}} > 0$ throughout the main paper for the cleanest visualizations and tightest radial bounds, but the underlying inductive bias is geometric, not enforced.

\section{Full Results}
\label{sec:full_results}
The complete per-horizon forecasting results for all 11 benchmark datasets across prediction lengths $H \in \{96, 192, 336, 720\}$ are presented in Table~\ref{tab:main_results}. These detailed results supplement the averaged metrics reported in Table~\ref{tab:main_results_mean}, which aggregates performance across all four horizons to provide a concise summary of model performance. 


\begin{table*}[h]
\centering
\setlength{\tabcolsep}{4pt}
\renewcommand{\arraystretch}{1.1}
\caption{Long-term forecasting results across 11 datasets and 4 horizons ($H \in \{96, 192, 336, 720\}$). HF: HyperForecast (ours). The input length L is set to 720 across all models. Best results are marked in \textbf{\textcolor{red}{red-bold}}; second-best are \textbf{\textcolor{blue}{blue-bold}}.}
\label{tab:main_results}
\resizebox{\textwidth}{!}{%
\begin{tabular}{l | c | 
cc | cc | cc | cc | cc | cc | cc | cc | cc | cc | cc | cc | cc | cc}
\toprule
\textbf{Methods} &  
& \multicolumn{2}{c}{HF NAR} 
& \multicolumn{2}{c}{HF AR}
& \multicolumn{2}{c}{PhaseFormer}
& \multicolumn{2}{c}{TimeBase}
& \multicolumn{2}{c}{SparseTSF}
& \multicolumn{2}{c}{TimeMixer}
& \multicolumn{2}{c}{FITS}
& \multicolumn{2}{c}{iTransformer}
& \multicolumn{2}{c}{DLinear}
& \multicolumn{2}{c}{PatchTST}
& \multicolumn{2}{c}{TimesNet}
& \multicolumn{2}{c}{FEDformer}
& \multicolumn{2}{c}{Autoformer}
& \multicolumn{2}{c}{Informer} \\
\cmidrule(lr){3-4}\cmidrule(lr){5-6}\cmidrule(lr){7-8}\cmidrule(lr){9-10}\cmidrule(lr){11-12}\cmidrule(lr){13-14}\cmidrule(lr){15-16}\cmidrule(lr){17-18}\cmidrule(lr){19-20}\cmidrule(lr){21-22}\cmidrule(lr){23-24}\cmidrule(lr){25-26}\cmidrule(lr){27-28}\cmidrule{29-30}
\textbf{Metrics}
 & \textbf{H} & 
 MSE & MAE
 & MSE & MAE 
 & MSE & MAE 
 & MSE & MAE 
 & MSE & MAE 
 & MSE & MAE 
 & MSE & MAE 
 & MSE & MAE 
 & MSE & MAE 
 & MSE & MAE 
 & MSE & MAE 
 & MSE & MAE 
 & MSE & MAE 
 & MSE & MAE \\

\midrule

\multirow{4}{*}{Weather}
& 96 
& \textbf{\textcolor{blue}{0.145}} & 0.206
& \textbf{\textcolor{red}{0.143}} & 0.202
& 0.148 & \textbf{\textcolor{red}{0.195}}
& 0.146 & \textbf{\textcolor{blue}{0.198}}
& 0.174 & 0.231
& 0.163 & 0.223
& 0.176 & 0.232
& 0.159 & 0.212
& 0.174 & 0.242
& 0.149 & 0.199
& 0.176 & 0.234
& 0.289 & 0.342
& 0.323 & 0.389
& 0.283 & 0.361 \\
& 192 
& \textbf{\textcolor{blue}{0.188}} & 0.252
& 0.190 & 0.248
& 0.193 & \textbf{\textcolor{red}{0.237}}
& \textbf{\textcolor{red}{0.185}} & \textbf{\textcolor{blue}{0.241}}
& 0.216 & 0.267
& 0.201 & 0.254
& 0.216 & 0.268
& 0.203 & 0.252
& 0.215 & 0.277
& 0.193 & 0.243
& 0.219 & 0.270
& 0.340 & 0.394
& 0.389 & 0.423
& 0.445 & 0.461 \\
& 336 
& \textbf{\textcolor{blue}{0.237}} & 0.292
& 0.238 & 0.289
& 0.242 & \textbf{\textcolor{red}{0.278}}
& \textbf{\textcolor{red}{0.236}} & \textbf{\textcolor{blue}{0.281}}
& 0.260 & 0.299
& 0.258 & 0.300
& 0.261 & 0.299
& 0.253 & 0.291
& 0.262 & 0.319
& 0.240 & 0.281
& 0.277 & 0.311
& 0.370 & 0.408
& 0.497 & 0.495
& 0.587 & 0.526 \\
& 720 
& \textbf{\textcolor{red}{0.303}} & 0.344
& 0.315 & 0.341
& 0.309 & \textbf{\textcolor{blue}{0.332}}
& \textbf{\textcolor{blue}{0.309}} & \textbf{\textcolor{red}{0.331}}
& 0.325 & 0.345
& 0.329 & 0.348
& 0.325 & 0.346
& 0.317 & 0.337
& 0.319 & 0.359
& 0.312 & 0.334
& 0.344 & 0.356
& 0.420 & 0.421
& 0.573 & 0.520
& 0.953 & 0.703 \\

\midrule

\multirow{4}{*}{AQShunyi}
& 96 
& \textbf{\textcolor{blue}{0.607}} & \textbf{\textcolor{blue}{0.475}}
& \textbf{\textcolor{red}{0.606}} & \textbf{\textcolor{red}{0.473}}
& 0.650 & 0.493
& 0.629 & 0.489
& 0.751 & 0.553
& 0.703 & 0.534
& 0.759 & 0.555
& 0.686 & 0.508
& 0.751 & 0.579
& 0.643 & 0.476
& 0.759 & 0.560
& 1.247 & 0.819
& 1.394 & 0.931
& 1.221 & 0.864 \\
& 192 
& \textbf{\textcolor{red}{0.640}} & \textbf{\textcolor{blue}{0.498}}
& \textbf{\textcolor{blue}{0.642}} & \textbf{\textcolor{red}{0.493}}
& 0.729 & 0.519
& 0.661 & 0.500
& 0.774 & 0.551
& 0.720 & 0.524
& 0.774 & 0.553
& 0.728 & 0.520
& 0.771 & 0.571
& 0.692 & 0.501
& 0.785 & 0.557
& 1.218 & 0.813
& 1.394 & 0.872
& 1.595 & 0.951 \\
& 336 
& \textbf{\textcolor{blue}{0.670}} & \textbf{\textcolor{blue}{0.510}}
& \textbf{\textcolor{red}{0.669}} & \textbf{\textcolor{red}{0.508}}
& 0.740 & 0.533
& 0.683 & 0.510
& 0.752 & 0.544
& 0.747 & 0.546
& 0.755 & 0.544
& 0.732 & 0.529
& 0.758 & 0.580
& 0.695 & 0.514
& 0.802 & 0.566
& 1.071 & 0.742
& 1.438 & 0.900
& 1.699 & 0.957 \\
& 720 
& \textbf{\textcolor{red}{0.712}} & 0.528
& 0.742 & 0.542
& 0.770 & 0.546
& \textbf{\textcolor{blue}{0.725}} & \textbf{\textcolor{blue}{0.523}}
& 0.763 & 0.537
& 0.772 & 0.541
& 0.763 & 0.538
& 0.744 & 0.524
& 0.748 & 0.559
& 0.732 & \textbf{\textcolor{red}{0.520}}
& 0.807 & 0.554
& 0.985 & 0.655
& 1.344 & 0.809
& 2.236 & 1.094 \\

\midrule

\multirow{4}{*}{AQWan}
& 96 
& \textbf{\textcolor{red}{0.692}} & \textbf{\textcolor{red}{0.458}}
& \textbf{\textcolor{blue}{0.696}} & \textbf{\textcolor{blue}{0.464}}
& 0.781 & 0.488
& 0.726 & 0.477
& 0.758 & 0.486
& 0.762 & 0.484
& 0.795 & 0.510
& 0.798 & 0.511
& 0.835 & 0.531
& 0.735 & 0.467
& 0.828 & 0.527
& 0.865 & 0.523
& 0.895 & 0.592
& 1.027 & 0.699 \\
& 192 
& \textbf{\textcolor{red}{0.733}} & \textbf{\textcolor{red}{0.479}}
& \textbf{\textcolor{blue}{0.742}} & 0.486
& 0.823 & 0.507
& 0.761 & 0.492
& 0.851 & 0.542
& 0.779 & 0.492
& 0.765 & \textbf{\textcolor{blue}{0.485}}
& 0.785 & 0.501
& 0.884 & 0.558
& 0.792 & 0.500
& 0.807 & 0.510
& 1.001 & 0.603
& 0.935 & 0.600
& 1.293 & 0.834 \\
& 336 
& \textbf{\textcolor{red}{0.760}} & \textbf{\textcolor{red}{0.492}}
& \textbf{\textcolor{blue}{0.766}} & \textbf{\textcolor{blue}{0.499}}
& 0.842 & 0.527
& 0.783 & 0.502
& 0.810 & 0.531
& 0.857 & 0.559
& 0.861 & 0.564
& 0.818 & 0.537
& 0.842 & 0.550
& 0.804 & 0.525
& 0.810 & 0.529
& 0.893 & 0.569
& 0.990 & 0.666
& 1.017 & 0.711 \\
& 720 
& \textbf{\textcolor{red}{0.819}} & \textbf{\textcolor{red}{0.514}}
& 0.846 & 0.532
& 0.900 & 0.547
& 0.847 & 0.524
& 0.883 & 0.546
& 0.891 & 0.549
& \textbf{\textcolor{blue}{0.831}} & \textbf{\textcolor{blue}{0.517}}
& 0.971 & 0.604
& 0.972 & 0.600
& 0.909 & 0.561
& 0.982 & 0.606
& 0.967 & 0.566
& 1.185 & 0.759
& 1.156 & 0.731 \\

\midrule

\multirow{4}{*}{CzeLan}
& 96 
& \textbf{\textcolor{red}{0.171}} & 0.236
& \textbf{\textcolor{blue}{0.172}} & 0.235
& 0.175 & \textbf{\textcolor{blue}{0.229}}
& 0.177 & \textbf{\textcolor{red}{0.224}}
& 0.204 & 0.263
& 0.182 & 0.233
& 0.208 & 0.268
& 0.202 & 0.259
& 0.187 & 0.239
& 0.184 & 0.236
& 0.204 & 0.262
& 0.208 & 0.260
& 0.251 & 0.334
& 0.293 & 0.384 \\
& 192 
& \textbf{\textcolor{red}{0.205}} & 0.267
& 0.212 & 0.278
& 0.219 & 0.271
& \textbf{\textcolor{blue}{0.211}} & \textbf{\textcolor{red}{0.250}}
& 0.227 & 0.274
& 0.216 & 0.259
& 0.242 & 0.292
& 0.233 & 0.282
& 0.226 & 0.271
& 0.214 & \textbf{\textcolor{blue}{0.257}}
& 0.256 & 0.308
& 0.258 & 0.307
& 0.285 & 0.360
& 0.329 & 0.399 \\
& 336 
& 0.237 & 0.305
& \textbf{\textcolor{blue}{0.234}} & 0.298
& 0.237 & \textbf{\textcolor{blue}{0.280}}
& 0.243 & \textbf{\textcolor{red}{0.276}}
& 0.239 & 0.305
& 0.246 & 0.312
& 0.247 & 0.316
& 0.239 & 0.305
& 0.238 & 0.301
& \textbf{\textcolor{red}{0.229}} & 0.290
& 0.274 & 0.347
& 0.299 & 0.365
& 0.263 & 0.354
& 0.366 & 0.476 \\
& 720 
& \textbf{\textcolor{blue}{0.270}} & 0.336
& 0.278 & 0.334
& 0.270 & \textbf{\textcolor{blue}{0.310}}
& \textbf{\textcolor{red}{0.269}} & \textbf{\textcolor{red}{0.297}}
& 0.292 & 0.325
& 0.291 & 0.322
& 0.303 & 0.338
& 0.305 & 0.340
& 0.315 & 0.349
& 0.288 & 0.319
& 0.299 & 0.332
& 0.300 & 0.324
& 0.357 & 0.423
& 0.399 & 0.446 \\

\midrule

\multirow{4}{*}{ZafNoo}
& 96 
& 0.418 & \textbf{\textcolor{blue}{0.399}}
& 0.420 & \textbf{\textcolor{red}{0.398}}
& 0.457 & 0.410
& 0.435 & 0.407
& 0.435 & 0.439
& \textbf{\textcolor{red}{0.410}} & 0.411
& 0.458 & 0.461
& 0.442 & 0.447
& 0.423 & 0.423
& \textbf{\textcolor{blue}{0.412}} & 0.413
& 0.443 & 0.444
& 0.484 & 0.473
& 0.548 & 0.566
& 0.651 & 0.672 \\
& 192 
& \textbf{\textcolor{blue}{0.469}} & 0.437
& \textbf{\textcolor{red}{0.466}} & \textbf{\textcolor{red}{0.431}}
& 0.524 & 0.450
& 0.470 & \textbf{\textcolor{blue}{0.435}}
& 0.548 & 0.509
& 0.473 & 0.437
& 0.506 & 0.468
& 0.560 & 0.522
& 0.534 & 0.494
& 0.518 & 0.479
& 0.572 & 0.529
& 0.579 & 0.512
& 0.738 & 0.712
& 0.745 & 0.730 \\
& 336 
& \textbf{\textcolor{blue}{0.508}} & \textbf{\textcolor{blue}{0.462}}
& \textbf{\textcolor{red}{0.507}} & 0.462
& 0.556 & 0.473
& 0.511 & \textbf{\textcolor{red}{0.453}}
& 0.525 & 0.474
& 0.576 & 0.520
& 0.528 & 0.477
& 0.536 & 0.486
& 0.596 & 0.537
& 0.528 & 0.477
& 0.579 & 0.523
& 0.566 & 0.499
& 0.624 & 0.564
& 0.662 & 0.601 \\
& 720 
& \textbf{\textcolor{red}{0.545}} & \textbf{\textcolor{red}{0.473}}
& 0.563 & 0.500
& 0.594 & 0.491
& 0.565 & \textbf{\textcolor{blue}{0.484}}
& 0.617 & 0.540
& \textbf{\textcolor{blue}{0.560}} & 0.487
& 0.590 & 0.515
& 0.623 & 0.546
& 0.606 & 0.527
& 0.581 & 0.506
& 0.587 & 0.511
& 0.776 & 0.671
& 0.758 & 0.698
& 0.847 & 0.783 \\

\midrule

\multirow{4}{*}{Exchange}
& 96 
& \textbf{\textcolor{red}{0.079}} & \textbf{\textcolor{red}{0.207}}
& \textbf{\textcolor{blue}{0.082}} & 0.216
& 0.107 & 0.236
& 0.087 & \textbf{\textcolor{blue}{0.207}}
& 0.083 & 0.211
& 0.090 & 0.229
& 0.094 & 0.241
& 0.096 & 0.244
& 0.096 & 0.243
& 0.083 & 0.211
& 0.088 & 0.224
& 0.092 & 0.224
& 0.119 & 0.316
& 0.106 & 0.276 \\
& 192 
& \textbf{\textcolor{blue}{0.165}} & 0.321
& \textbf{\textcolor{red}{0.159}} & \textbf{\textcolor{red}{0.303}}
& 0.173 & \textbf{\textcolor{blue}{0.303}}
& 0.185 & 0.306
& 0.172 & 0.317
& 0.167 & 0.305
& 0.167 & 0.307
& 0.176 & 0.325
& 0.169 & 0.310
& 0.168 & 0.307
& 0.173 & 0.317
& 0.208 & 0.355
& 0.237 & 0.444
& 0.252 & 0.482 \\
& 336 
& \textbf{\textcolor{blue}{0.305}} & 0.457
& \textbf{\textcolor{red}{0.263}} & \textbf{\textcolor{red}{0.395}}
& 0.313 & 0.405
& 0.355 & 0.436
& 0.308 & \textbf{\textcolor{blue}{0.395}}
& 0.342 & 0.437
& 0.319 & 0.409
& 0.312 & 0.399
& 0.344 & 0.438
& 0.324 & 0.413
& 0.362 & 0.462
& 0.405 & 0.512
& 0.437 & 0.561
& 0.536 & 0.700 \\
& 720 
& \textbf{\textcolor{blue}{0.607}} & 0.647
& \textbf{\textcolor{red}{0.591}} & \textbf{\textcolor{red}{0.602}}
& 1.423 & 0.863
& 0.608 & \textbf{\textcolor{blue}{0.619}}
& 0.701 & 0.716
& 0.721 & 0.734
& 0.629 & 0.643
& 0.619 & 0.635
& 0.637 & 0.648
& 0.659 & 0.670
& 0.683 & 0.696
& 0.813 & 0.795
& 0.994 & 1.035
& 0.835 & 0.865 \\

\midrule

\multirow{4}{*}{METR-LA}
& 96 
& 1.038 & 0.702
& 1.053 & \textbf{\textcolor{red}{0.635}}
& 1.120 & 0.657
& \textbf{\textcolor{blue}{1.028}} & 0.655
& 1.056 & 0.667
& 1.051 & 0.657
& 1.047 & 0.661
& \textbf{\textcolor{red}{1.023}} & \textbf{\textcolor{blue}{0.640}}
& 1.096 & 0.685
& 1.053 & 0.658
& 1.206 & 0.755
& 1.440 & 0.882
& 1.443 & 0.908
& 1.666 & 1.088 \\
& 192 
& 1.239 & 0.706
& 1.248 & 0.759
& 1.264 & 0.732
& \textbf{\textcolor{red}{1.112}} & 0.711
& 1.232 & 0.730
& 1.163 & \textbf{\textcolor{blue}{0.689}}
& 1.295 & 0.772
& 1.196 & 0.713
& 1.330 & 0.788
& \textbf{\textcolor{blue}{1.134}} & \textbf{\textcolor{red}{0.672}}
& 1.359 & 0.806
& 1.478 & 0.839
& 1.366 & 0.830
& 1.452 & 0.904 \\
& 336 
& 1.324 & 0.743
& 1.354 & 0.786
& 1.398 & 0.795
& 1.317 & 0.771
& 1.356 & 0.772
& \textbf{\textcolor{blue}{1.304}} & \textbf{\textcolor{blue}{0.737}}
& 1.309 & 0.743
& 1.320 & 0.750
& 1.353 & 0.764
& \textbf{\textcolor{red}{1.250}} & \textbf{\textcolor{red}{0.706}}
& 1.376 & 0.778
& 1.555 & 0.860
& 1.525 & 0.883
& 2.036 & 1.209 \\
& 720 
& 1.462 & 0.812
& 1.528 & 0.837
& 1.611 & 0.864
& 1.558 & 0.877
& 1.562 & 0.885
& 1.564 & 0.879
& 1.534 & 0.865
& 1.474 & 0.834
& \textbf{\textcolor{blue}{1.423}} & \textbf{\textcolor{blue}{0.800}}
& \textbf{\textcolor{red}{1.404}} & \textbf{\textcolor{red}{0.789}}
& 1.619 & 0.910
& 1.725 & 0.921
& 2.221 & 1.277
& 2.420 & 1.384 \\

\midrule

\multirow{4}{*}{PM2.5}
& 96 
& \textbf{\textcolor{red}{0.416}} & \textbf{\textcolor{blue}{0.437}}
& \textbf{\textcolor{blue}{0.416}} & 0.437
& 0.435 & 0.438
& 0.432 & \textbf{\textcolor{red}{0.435}}
& 0.475 & 0.482
& 0.445 & 0.449
& 0.495 & 0.501
& 0.466 & 0.474
& 0.437 & 0.441
& 0.450 & 0.454
& 0.462 & 0.467
& 0.560 & 0.550
& 0.596 & 0.615
& 0.566 & 0.593 \\
& 192 
& \textbf{\textcolor{red}{0.418}} & 0.444
& \textbf{\textcolor{blue}{0.418}} & 0.441
& 0.440 & \textbf{\textcolor{blue}{0.435}}
& 0.429 & \textbf{\textcolor{red}{0.432}}
& 0.458 & 0.465
& 0.479 & 0.483
& 0.455 & 0.460
& 0.505 & 0.511
& 0.454 & 0.458
& 0.461 & 0.465
& 0.452 & 0.456
& 0.480 & 0.474
& 0.567 & 0.614
& 0.670 & 0.678 \\
& 336 
& \textbf{\textcolor{red}{0.406}} & 0.449
& \textbf{\textcolor{blue}{0.416}} & 0.443
& 0.428 & \textbf{\textcolor{red}{0.430}}
& 0.426 & \textbf{\textcolor{blue}{0.433}}
& 0.478 & 0.482
& 0.473 & 0.475
& 0.450 & 0.453
& 0.461 & 0.467
& 0.449 & 0.451
& 0.475 & 0.477
& 0.490 & 0.492
& 0.586 & 0.555
& 0.671 & 0.684
& 0.749 & 0.793 \\
& 720 
& \textbf{\textcolor{blue}{0.400}} & 0.477
& \textbf{\textcolor{red}{0.374}} & \textbf{\textcolor{red}{0.427}}
& 0.418 & \textbf{\textcolor{blue}{0.433}}
& 0.425 & 0.448
& 0.456 & 0.479
& 0.464 & 0.487
& 0.440 & 0.461
& 0.489 & 0.514
& 0.457 & 0.479
& 0.448 & 0.470
& 0.461 & 0.484
& 0.612 & 0.609
& 0.652 & 0.686
& 0.561 & 0.589 \\

\midrule

\multirow{4}{*}{Solar}
& 96 
& 0.177 & 0.246
& \textbf{\textcolor{red}{0.170}} & \textbf{\textcolor{blue}{0.227}}
& \textbf{\textcolor{blue}{0.176}} & \textbf{\textcolor{red}{0.226}}
& 0.192 & 0.239
& 0.205 & 0.241
& 0.232 & 0.271
& 0.218 & 0.257
& 0.217 & 0.255
& 0.200 & 0.234
& 0.205 & 0.239
& 0.243 & 0.284
& 0.231 & 0.254
& 0.279 & 0.339
& 0.308 & 0.377 \\
& 192 
& \textbf{\textcolor{blue}{0.204}} & 0.281
& 0.206 & 0.256
& \textbf{\textcolor{red}{0.194}} & \textbf{\textcolor{red}{0.237}}
& 0.213 & \textbf{\textcolor{blue}{0.252}}
& 0.215 & 0.265
& 0.238 & 0.293
& 0.207 & 0.256
& 0.208 & 0.257
& 0.239 & 0.294
& 0.227 & 0.280
& 0.253 & 0.311
& 0.257 & 0.301
& 0.337 & 0.418
& 0.332 & 0.422 \\
& 336 
& 0.221 & 0.289
& 0.227 & 0.282
& \textbf{\textcolor{red}{0.199}} & \textbf{\textcolor{red}{0.247}}
& 0.222 & \textbf{\textcolor{blue}{0.261}}
& \textbf{\textcolor{blue}{0.213}} & 0.276
& 0.234 & 0.301
& 0.229 & 0.295
& 0.238 & 0.309
& 0.231 & 0.298
& 0.225 & 0.290
& 0.222 & 0.287
& 0.260 & 0.319
& 0.263 & 0.352
& 0.314 & 0.421 \\
& 720 
& \textbf{\textcolor{blue}{0.217}} & 0.280
& 0.243 & 0.306
& \textbf{\textcolor{red}{0.203}} & \textbf{\textcolor{red}{0.250}}
& 0.235 & \textbf{\textcolor{blue}{0.264}}
& 0.232 & 0.272
& 0.273 & 0.319
& 0.261 & 0.307
& 0.270 & 0.319
& 0.237 & 0.277
& 0.249 & 0.291
& 0.254 & 0.298
& 0.260 & 0.290
& 0.345 & 0.413
& 0.355 & 0.424 \\

\midrule

\multirow{4}{*}{Temp}
& 96 
& \textbf{\textcolor{blue}{0.123}} & \textbf{\textcolor{blue}{0.270}}
& \textbf{\textcolor{red}{0.121}} & \textbf{\textcolor{red}{0.266}}
& 0.147 & 0.294
& 0.142 & 0.286
& 0.149 & 0.300
& 0.158 & 0.316
& 0.168 & 0.338
& 0.155 & 0.310
& 0.165 & 0.329
& 0.145 & 0.291
& 0.155 & 0.309
& 0.173 & 0.338
& 0.201 & 0.411
& 0.227 & 0.459 \\
& 192 
& \textbf{\textcolor{red}{0.123}} & \textbf{\textcolor{red}{0.272}}
& \textbf{\textcolor{blue}{0.125}} & \textbf{\textcolor{blue}{0.272}}
& 0.161 & 0.311
& 0.155 & 0.305
& 0.168 & 0.323
& 0.163 & 0.312
& 0.181 & 0.351
& 0.168 & 0.324
& 0.174 & 0.335
& 0.166 & 0.319
& 0.187 & 0.358
& 0.179 & 0.326
& 0.246 & 0.491
& 0.281 & 0.564 \\
& 336 
& \textbf{\textcolor{red}{0.117}} & \textbf{\textcolor{red}{0.265}}
& \textbf{\textcolor{blue}{0.132}} & \textbf{\textcolor{blue}{0.284}}
& 0.188 & 0.336
& 0.176 & 0.323
& 0.220 & 0.373
& 0.249 & 0.421
& 0.223 & 0.381
& 0.232 & 0.395
& 0.246 & 0.417
& 0.221 & 0.374
& 0.236 & 0.400
& 0.250 & 0.412
& 0.312 & 0.554
& 0.349 & 0.606 \\
& 720 
& \textbf{\textcolor{red}{0.113}} & \textbf{\textcolor{red}{0.260}}
& \textbf{\textcolor{blue}{0.203}} & \textbf{\textcolor{blue}{0.352}}
& 0.371 & 0.451
& 0.274 & 0.437
& 0.642 & 0.654
& 0.637 & 0.643
& 0.689 & 0.699
& 0.652 & 0.663
& 0.584 & 0.590
& 0.632 & 0.639
& 0.636 & 0.642
& 0.769 & 0.747
& 0.859 & 0.893
& 1.026 & 1.101 \\

\midrule

\multirow{4}{*}{Wind}
& 96 
& \textbf{\textcolor{blue}{0.771}} & 0.618
& \textbf{\textcolor{red}{0.765}} & 0.616
& 0.824 & 0.653
& 0.785 & 0.626
& 0.796 & \textbf{\textcolor{blue}{0.589}}
& 0.908 & 0.669
& 0.787 & \textbf{\textcolor{red}{0.581}}
& 0.824 & 0.609
& 0.824 & 0.607
& 0.802 & 0.591
& 0.892 & 0.657
& 1.008 & 0.718
& 1.110 & 0.847
& 1.132 & 0.836 \\
& 192 
& \textbf{\textcolor{red}{0.893}} & \textbf{\textcolor{blue}{0.679}}
& \textbf{\textcolor{blue}{0.897}} & 0.683
& 1.001 & 0.728
& 0.902 & 0.690
& 0.949 & 0.698
& 1.040 & 0.764
& 0.910 & \textbf{\textcolor{red}{0.674}}
& 1.054 & 0.779
& 0.961 & 0.707
& 0.944 & 0.694
& 1.008 & 0.741
& 1.256 & 0.876
& 1.218 & 0.954
& 1.540 & 1.188 \\
& 336 
& \textbf{\textcolor{blue}{1.011}} & \textbf{\textcolor{red}{0.734}}
& 1.015 & 0.748
& 1.090 & 0.774
& \textbf{\textcolor{red}{0.997}} & 0.737
& 1.182 & 0.851
& 1.039 & 0.747
& 1.100 & 0.795
& 1.025 & 0.742
& 1.022 & \textbf{\textcolor{blue}{0.734}}
& 1.043 & 0.750
& 1.185 & 0.853
& 1.375 & 0.923
& 1.209 & 0.935
& 1.399 & 1.015 \\
& 720 
& \textbf{\textcolor{red}{1.045}} & \textbf{\textcolor{red}{0.768}}
& 1.121 & 0.805
& 1.157 & 0.816
& \textbf{\textcolor{blue}{1.075}} & 0.782
& 1.127 & 0.808
& 1.189 & 0.851
& 1.295 & 0.932
& 1.196 & 0.863
& 1.266 & 0.907
& 1.084 & \textbf{\textcolor{blue}{0.776}}
& 1.320 & 0.946
& 1.281 & 0.846
& 1.410 & 1.053
& 1.453 & 1.094 \\

\bottomrule
\end{tabular}
}
\end{table*}

\section{Implementation Details}
\label{app:implementation_details}
All experiments were conducted on a single NVIDIA A100 GPU (40GB) with CUDA 12.6.0. We implemented HyperForecast in PyTorch~\citep{paszke2019pytorch} with GeoOpt~\citep{kochurov2020geoopt} for Riemannian optimization on hyperbolic manifolds. The model was trained for 30 epochs using the RiemannianAdam optimizer with OneCycleLR scheduling~\citep{smith2019super}. We set the embedding dimension to 64 and hidden dimension to 256 across all experiments. The hyperbolic curvature was initialized to 1.0 with learnable scaling parameters for exponential map stability. Gradient clipping with a maximum norm of 5.0 was applied to prevent numerical instability in hyperbolic operations. We used a batch size of 32 for all datasets except METR-LA (207 channels, the dataset with the highest dimensionality in our benchmark suite). Dropout rates of 0.3 for encoding layers and 0.2 for reconstruction layers were employed for regularization. We performed hyperparameter tuning via grid search, exploring temporal consistency loss weights $\{0.1, 0.2\}$, hierarchy loss weights $\{1 \times 10^{-4}, 1 \times 10^{-3}, 0.1\}$, learning rates $\{1 \times 10^{-1}, 1 \times 10^{-2}, 1 \times 10^{-3}\}$, and moving window sizes $\{2, 4, 5, 6\}$. The best-performing configurations were selected based on validation MSE for each prediction horizon. RevIN normalization~\citep{kim2022reversible} was applied to all inputs. We used early stopping with a patience of 5 epochs, monitored on validation loss. All models used a lookback window of 720 timesteps, prediction horizons of $\{96, 192, 336, 720\}$, and contiguous 7:1:2 train/validation/test splits, following standard LTSF benchmarks~\citep{wu2021autoformer,huang2025timebase}.

\section{Code Availability}
\label{app:code_avail}
The code is available at an anonymized GitHub repository: \url{https://anonymous.4open.science/r/Hyperbolic_TimeSeries-3876/}.
\end{document}

%% file: math_commands.tex
\usepackage{amsmath,amsfonts,bm}

\def\eqref#1{equation~\ref{#1}}

\def\1{\bm{1}}

\DeclareMathAlphabet{\mathsfit}{\encodingdefault}{\sfdefault}{m}{sl}
\SetMathAlphabet{\mathsfit}{bold}{\encodingdefault}{\sfdefault}{bx}{n}

